\documentclass{ieeetj}
\usepackage{times}
\usepackage{setspace}
\usepackage{url}
\spacing{1}
\usepackage[utf8]{inputenc}
\usepackage[T1]{fontenc}
\usepackage{graphicx}		
\usepackage{wrapfig}
\usepackage{sidecap} 
\usepackage[export]{adjustbox}
\usepackage{float}

\usepackage{comment}

\usepackage{xcolor}

\usepackage{amsmath} 
\usepackage{amssymb}  
\usepackage{amsthm}
\usepackage{mathtools}
\usepackage[normalem]{ulem}
\usepackage{paralist}	
\usepackage[space]{grffile} 
\usepackage{color}
\let\labelindent\relax
\usepackage{enumitem}
\usepackage{bm}
\usepackage{cancel}
\usepackage{hhline}
\usepackage[c2 , nocomma]{optidef}
\usepackage{oubraces}
\usepackage{algorithmic}
\usepackage{graphicx}
\usepackage{textcomp}

\usepackage{moreverb, url}

\usepackage[ruled,vlined]{algorithm2e}
\usepackage{stackengine}
\usepackage{dirtytalk}

\usepackage{array}
\usepackage{verbatim}
\usepackage{upgreek}

\usepackage{amsfonts}
\usepackage{lipsum}
\usepackage{cite}
\usepackage{hyperref}
\hypersetup{hidelinks=true}

\theoremstyle{definition}

\theoremstyle{remark}
\newtheorem{remark}{Remark}
\theoremstyle{definition}
\theoremstyle{definition}

\DeclareMathOperator*{\argmax}{arg\,max}
\DeclareMathOperator*{\argmin}{arg\,min}

\definecolor{blue}{RGB}{38,38,134}
\definecolor{darkblue}{RGB}{0,0,102}
\definecolor{lightblue}{RGB}{77,77,148}

\definecolor{gold}{RGB}{234, 170, 0}
\definecolor{metallic_gold}{RGB}{139, 111, 78}

\DeclareMathOperator{\rank}{rank}

\usepackage{xfrac}

\usepackage{dblfloatfix}

\usepackage{pdfpages}
\usepackage{graphicx}

\def\IEEEbibitemsep{0pt plus .5pt}

\usepackage{enumitem}

\def\BibTeX{{\rm B\kern-.05em{\sc i\kern-.025em b}\kern-.08em
    T\kern-.1667em\lower.7ex\hbox{E}\kern-.125emX}}
\AtBeginDocument{\definecolor{tmlcncolor}{cmyk}{0.93,0.59,0.15,0.02}\definecolor{NavyBlue}{RGB}{0,86,125}}

\def\OJlogo{\vspace{-4pt}$<$logo$>$}
\def\seclogo{\vspace{10pt}$<$logo$>$}

\def\authorrefmark#1{\ensuremath{^{\textbf{#1}}}}

\begin{document}
\receiveddate{23 February, 2025}
\reviseddate{28 May, 2025}
\accepteddate{02 July, 2025}
\publisheddate{XX Month, XXXX}
\currentdate{XX Month, XXXX}
\doiinfo{XXXX.2022.1234567}

\markboth{}{Touma*, Da\c{s}* {et al.}}

\title{ \bf AI Space Cortex: An Experimental System for Future Era Space Exploration}

\author{
Thomas Touma$^{*}$\authorrefmark{1, 5},
Ersin Da\c{s}$^{*}$\authorrefmark{1},
Erica Tevere\authorrefmark{2}, \\
Martin Feather\authorrefmark{2},
Ksenia Kolcio\authorrefmark{3},
Maurice Prather\authorrefmark{3}, \\
Alberto Candela\authorrefmark{2},
Ashish Goel\authorrefmark{2},
Erik Kramer\authorrefmark{4}, \\
Hari Nayar\authorrefmark{2},
Lorraine Fesq\authorrefmark{2},
Joel W. Burdick\authorrefmark{1}%
\affil{T. Touma, E. Da\c{s}, and J. W. Burdick are with the Department of Mechanical and Civil Engineering, 
California Institute of Technology, Pasadena, CA 91125, USA.
{\tt\small \{ttouma, ersindas, jburdick\}@caltech.edu}}%
\affil{E. Tevere, M. Feather, A. Candela, A. Goel, H. Nayar, and L. Fesq are with the Jet Propulsion Laboratory, California Institute of Technology, Pasadena, CA 91109, USA. {\tt\small \{erica.l.tevere, martin.s.feather, alberto.candela.garza, ashish.goel, hdnayar, lorraine.m.fesq\}@jpl.nasa.gov}}%
\affil{K. Kolcio and M. Prather are with Okean Solutions, Inc.,
1211 E. Denny Way, \#32A
Seattle, WA 98112, USA. {\tt\small \{ksenia, maurice\}@okean.solutions}
}%
\affil{E. Kramer is with the University of California, Los Angeles, CA 90095, USA. {\tt\small \{ehkramer\}@g.ucla.edu}}%
\affil{T. Touma 
is also with Stealth Labs, Pasadena, CA 91106, USA. {\tt\small \{thomas\}@stealthlabshq.com}}%
}
\corresp{Corresponding authors: Thomas Touma (email: thomas@stealthlabshq.com), Ersin Da\c{s} (email: ersindas@caltech.edu).}
\authornote{*This work was supported by NASA Grant 80NSSC21K1032. A portion of this research was carried out at the Jet Propulsion Laboratory, California Institute of Technology, under a contract with the National Aeronautics and Space Administration (80NM0018D0004). \\
*Both authors contributed equally.}%

\begin{abstract} 
Our Robust, Explainable Autonomy for Scientific Icy Moon Operations (REASIMO) effort contributes to NASA’s Concepts for Ocean worlds Life Detection Technology (COLDTech) program, which explores science platform technologies for ocean worlds such as Europa and Enceladus. Ocean world missions pose significant operational challenges. These include long communication lags, limited power, and lifetime limitations caused by radiation damage and hostile conditions. Given these operational limitations, onboard autonomy will be vital for future Ocean world missions. Besides the management of nominal lander operations, onboard autonomy must react appropriately in the event of anomalies. Traditional spacecraft rely on a transition into 'safe-mode' in which non-essential components and subsystems are powered off to preserve safety and maintain communication with Earth. For a severely time-limited Ocean world mission, resolutions to these anomalies that can be executed without Earth-in-the-loop communication and associated delays are paramount for completion of the mission objectives and science goals. To address these challenges, the REASIMO effort aims to demonstrate a robust level of AI-assisted autonomy for such missions, including the ability to detect and recover from anomalies, and to perform missions based on pre-trained behaviors rather than hard-coded, predetermined logic like all prior space missions. We developed an AI-assisted, personality-driven, intelligent framework for control of an Ocean world mission by combining a mix of advanced technologies. To demonstrate the capabilities of the framework, we perform tests of autonomous sampling operations on a lander-manipulator testbed at the NASA Jet Propulsion Laboratory, approximating possible surface conditions such a mission might encounter. This paper presents the architecture and encapsulated functionality of our intelligent mission control framework, \textit{AI Space Cortex}, and reports results from deployment of this technology on a flight-relevant testbed to demonstrate its handling of the challenges of autonomous sampling operations across changing system conditions. 
\end{abstract}

\begin{IEEEkeywords}
 Fault Detection, Icy Moon Operations, Kinematic Calibration, Large Language Models (LLM), Planetary Robotics, Segmentation, Space Exploration, Vision Foundation Models.
\end{IEEEkeywords}

\maketitle

\section{INTRODUCTION} 
\label{sec:intro}
\begin{figure*}[t] 
    \centering
    \includegraphics[width=1\linewidth]{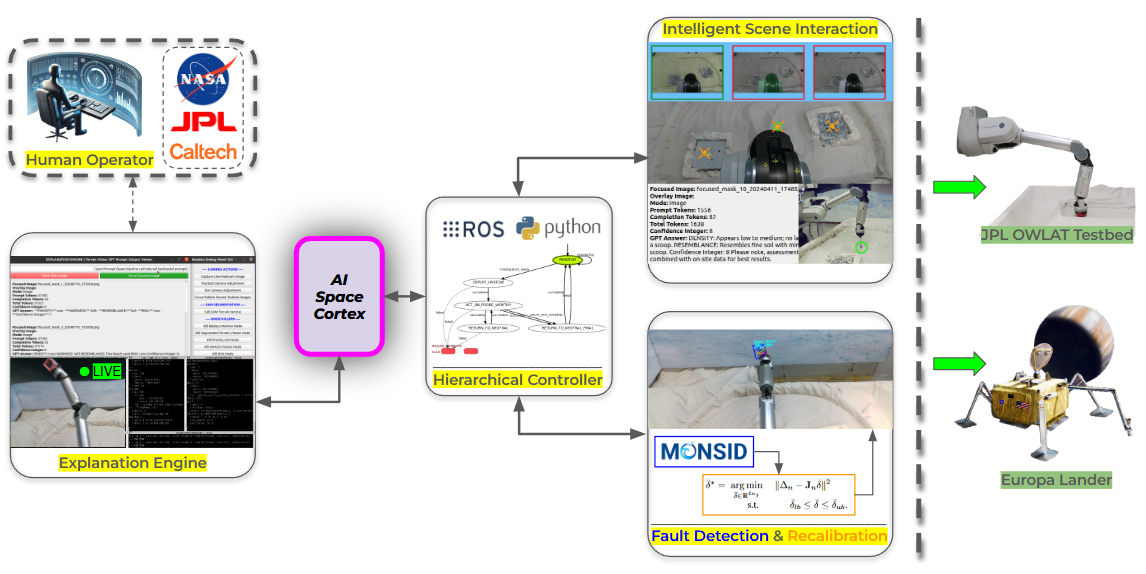}
    \caption{Overview of the AI Space Cortex, developed as part of the Robust, Explainable Autonomy for Scientific Icy Moon Operations (REASIMO) effort, and funded by NASA's Concepts for Ocean worlds Life Detection Technology (COLDTech) program. Its various components will be discussed in detail throughout the paper.}
    \label{fig:main-figure}
\end{figure*}

\IEEEPARstart{N}{ASA's} quest to discover extant life on habitable worlds in the Solar System has identified “ocean worlds” as leading candidates for exploration \cite{hendrix2019nasa}. Ocean worlds are those known or believed to have a current liquid ocean. In addition to Earth, these are Callisto, Enceladus, Europa, Ganymede, and Triton. Following its successful launch on October 14, 2024, the Europa Clipper is on its way to assess the habitability of Europa, a moon of Jupiter \cite{pappalardo2024science}. Europa Clipper (via highly eccentric orbits) will perform almost 50 close flybys of Europa itself, while largely out of Jupiter’s high radiation environment \cite{roberts2023exploring}. Results from this mission may spur the next step in exploration of these ocean worlds: in-situ investigation via a landed mission on Europa. Ideas for a Europa Lander mission have long been studied; \cite{hand2022science} presents the concept as developed in recent years. Such a landed mission would face key challenges of limited lifetime (post-landing) due to finite battery power, harsh environmental conditions (notably radiation), and long latency in communication with distant Earth, over 4 Astronomical Units (AU) away. Furthermore, the lander design may not include an orbiting communication relay. Instead, the lander would directly perform communication to Earth, whose availability is constrained to approximately 40 hours per 85-hour orbit of Europa. 

The Europa Lander concept represents the kind of future space missions that will require synergy between robotics and autonomy to push the boundaries of planetary exploration and mission management \cite{nesnas2021autonomy}. In these missions, not only must the on-board autonomy control the spacecraft's mechanisms (robotic arm to excavate and gather samples, in-situ sample analyzers, and communication of analysis results to Earth), but it must also make more complex decisions that traditionally require human-in-the-loop feedback or higher level knowledge of the system as a whole. Decisions like selecting among science-informed choices of sampling site, whether to gather additional samples from the same site based on indications of biosignatures, when to detect, diagnose, and triage anomalies and failures, and when to stop all activities to use the remaining energy solely to transmit data back to Earth. Demonstrations, both in software and on a hardware lander prototype, of existing autonomous software able to make these decisions with limited human intervention are covered in \cite{wagner2024}. 

For the design and development of an actual Lander, every attempt would be made to assure the correct operation of its hardware and software. This is traditionally checked by a rigorous verification and validation process (pre-launch). Nevertheless, the spacecraft is bound to experience failures over its mission lifetime, as a result of exposure to the harsh environment or due to off-nominal operating conditions. To guard against these the lander would need to be resilient, \cite{banerjee2023resiliency} able to autonomously recover from faults and failures and continue nominal operations when possible.

Our paper focuses on the development of an AI-assisted autonomy framework, \textit{AI Space Cortex}, that builds resiliency, management, personality, and pre-trained logic into its spacecraft operations. An overview of the AI Space Cortex can be seen in Fig.~\ref{fig:main-figure}. We are specifically interested in performing end-to-end surface sampling operations, which include sample site selection, successful sample collection, and sample deposit in science instruments. Because we focus on the resiliency of the autonomy framework, we introduce a range of anomalies into this scenario to be detected, diagnosed, and recovered from, all without the need for human intervention when possible. To achieve this goal, the AI Space Cortex framework combines (1) a state-machine based task-level executive for directing robot sampling and science system behavior; (2) the Model-based Off-Nominal State Identification and Detection (MONSID) system \cite{kolcio2016model} for fault detection and fault source isolation; (3) alternate recovery actions dictated by our executive and informed by MONSID’s situation assessment; (4) an AI based, intelligent vision system which informs based on its own reasoning where and when science sampling should take place; and (5) a graphical operator interface (Explanation Engine) used to inform operators of the robot's behavior. Recovery can be simple (e.g., retry, restart) or complex (e.g., automated recalibration of a damaged robot sampling arm). We implement these recovery capabilities as modules within the AI Space Cortex framework. We then illustrate the capabilities of this framework by demonstrating the sampling scenario and introducing possible anomalies on the COLDTech developed Ocean Worlds Lander Autonomy Testbed (OWLAT) - a representative hardware testbed for autonomous sampling operations \cite{tevere2024owlat}.

The rest of this paper is organized as follows. Section ~\ref{sec:cortex} introduces the architecture and functionality of AI Space Cortex, which is our high-level decision making framework used for mission decision making. Section~\ref{sec:hierarchical} provides details on the hierarchical controller used to execute tasks safely and efficiently. Section~\ref{sec:segmentation} describes our novel intelligent scene interaction module, which allows autonomous scene analysis and science site selection. Section~\ref{sec:explanation} summarizes the explanation engine, which provides intuitive interaction with and explanations of system actions. Section~\ref{sec:monsid} introduces the MONSID system, which detects and diagnoses system faults and determines when the system is operating in an off-nominal way. Section~\ref{sec:recal} introduces a novel module to perform non-parametric and data-efficient online recalibration as a form of fault recovery. Section~\ref{sec:owlat} introduces the hardware testbed used to perform experiments with the AI Space Cortex framework. Section~\ref{sec:results} presents the results of validation on system-level experiments. Finally, Section~\ref{sec:conclusion} presents our conclusions and future directions for this work.

\section{RELATED WORK}
\label{sec:lit_review}
In this paper, we present our AI-based autonomy framework, \textit{AI Space Cortex}, that aims to meld modern technology and practices into an autonomy framework for spacecraft surface operations. Autonomous spacecraft operations have been developed significantly in the past decades; they are critical pieces of technology to enable planetary missions that go further and do more. 

\subsection{End-to-End Surface-Sampling Frameworks}
In practice, autonomous operations are restricted in scope in spacecraft surface operations where performance and risk-management is key and computation is limited. Examples of autonomy being integrated into recent spacecraft operations include OSIRIS-REx \cite{lauretta2017osiris} and Mars Perseverance Rover \cite{verma2023autonomous}. In these cases some portion of operations are autonomous along some time-horizon, but the spacecraft still operates with humans-in-the-loop to plan and execute larger mission objectives and handle anomalous or particularly complex operational conditions. NASA's OSIRIS-REx mission collected and returned sample from Asteroid (101955) Bennu. The OSIRIS-REx spacecraft itself was single-fault tolerant and was equipped with autonomous fault detection, isolation, and recovery on-board. In addition, a portion of the sampling operations were completed autonomously. To deliver the Touch and Go Sample Acquisition Mechanism (TAGSAM) sampling head to within 25 m of a selected point on the surface the spacecraft had two independent autonomous guidance systems for the final closure with the asteroid surface during sample acquisition \cite{lauretta2017osiris}. While autonomy was used to execute a critical part of the sampling operations in this mission, sample site selection and sample collection verification both required interaction from the team on Earth.

For NASA's Perseverance rover, whose mission was to select, document, core, and deploy a high-value sample collection on the surface of Mars within one Mars year of landing has greatly benefited from onboard autonomy. Autonomous operations have become an integrated portion of daily driving, selecting and observing science targets, and onboard activity scheduling for the Perseverance rover, respectively enabled by the development of the autonomous navigation (AutoNav), Autonomous Exploration for Gathering Increased Science (AEGIS), and OnBoard Planner (OBP) capabilities. Each of these capabilities allows for optimized, resource-efficient surface operations, enabling the team to push to achieve the science mission objectives quickly \cite{verma2023autonomous}. While here autonomy is a greater portion of the sampling operations, the time-horizon is limited before ground-in-the-loop (GITL) interaction happens so that the team can react and respond on the next sol.

There is also a large existing body of work developing fully autonomous end-to-end surface operation pipelines. While many of these solutions have not been deployed on a flight mission they include works deployed on prototyped systems in Earth analog field test locations. Most notably is the work that came out of the Europa Lander Mission Concept as detailed in Section ~\ref{sec:intro}. This included the development on an autonomy prototype with a hierarchical utility model that is used to maximize both the amount of expected science return as well as the number of mission objectives imposed by the ground \cite{wagner2024}, \cite{bowkett2025}. The goal of this autonomy pipeline was to carry out complex, science-centric missions with limited interventions from humans, and this was shown on the prototyped hardware system in an operational readiness test on the Matanuska Glacier in Alaska. Our work, while looking at a very similar problem, end-to-end autonomous surface sampling operations for a lander-like mission, takes a different approach by leveraging modern tools in AI, like large language models, to autonomously initiate and execute science missions.

\subsection{Large-Model Reasoning for Scientific Decision-Making}
Scientific decision making is a difficult task often including teams of highly-trained scientist and an array of processed data inputs. With the growing ability of foundation models to perform complex tasks there has been an emerging interest in using foundation models, specifically vision language models (VLMs), for scientific reasoning tasks where representative data may be less common in training datasets. 

Some works benchmark how models perform on tasks to answer biologically relevant questions. In \cite{maruf2024vlm4bio}, pre-trained VLMs are evaluated on their ability to aid scientists in answering a range of organismal biology questions without any additional fine-tuning. The VLMs perform well on simple or well-constrained biologically relevant tasks but struggle in complex task settings that are practically more relevant to biologists.

Another area of interest is for Earth observation tasks such as scene understanding using satellite and aerial images \cite{danish2024geobench}, \cite{zhang2024good}. Concluding that state-of-the-art VLMs possess world knowledge that leads to strong performance in some tasks, but continue to have poor spatial reasoning capabilities, limiting their usefulness in others.

There has also been interest in applying these foundation models to space applications, given the inherent zero-shot nature of space exploration. \cite{foutter2024adapting} does this by fine-tuning an open-source VLM on an augmented dataset and benchmarking it against semantic reasoning tasks both in and out of distribution. While \cite{foutter2024adapting} uses GPT-assisted annotation to supplement existing extraterrestrial datasets, in this work, we use those annotations to perform semantic reasoning directly on the environment to prioritize science investigation targets. One notable difference in approach is that we ask the model to do semantic reasoning tasks on pre-segmented images, narrowing the scope of our semantic reasoning task and removing additional context whereas \cite{foutter2024adapting} uses full images.

\subsection{Fault Detection, Isolation, and Recovery}
In \cite{tipaldi2015survey}, the authors outline how spacecraft projects currently design, implement, and verify fault detection, isolation, and recovery across all mission phases, connecting industrial practices to hierarchies of hardware and software autonomy.

A fault detection, isolation, and recovery architecture that enables JAXA’s Martian Moons eXploration (MMX) Phobos rover to recognise and autonomously correct mobility faults is proposed in \cite{skibbe2023fault}. Built on an exhaustive failure analysis, the scheme ensures that the rover can maintain locomotion—even during critical landing and deployment phases—without ground intervention.

An online belief-space tree-search planner has been developed to actively select control actions that identify ambiguous actuator or sensor faults, all while ensuring safety in spacecraft motion under chance constraints \cite{ragan2024online}. Tests conducted on a free-floating spacecraft simulator, along with supporting simulations, demonstrate that this method isolates faults more rapidly and avoids collisions more reliably compared to passive or greedy baseline approaches.

\subsection{Kinematic Calibration}
Kinematic calibration of a robot manipulator has been extensively studied. Typically, the calibration process requires expert inputs for experiment design and numerous measurements to optimize an observability index. After data acquisition, offline optimization-based techniques, such as iterative least squares regression with linearized kinematic equations \cite{hollerbach1996}, Quadratic-Programming (QP) \cite{cursi2021} are used to identify the kinematic parameters. Although these approaches improve robot positioning accuracy, they are neither data-efficient or capable of completely autonomous implementation. Both characteristics are needed for a deep-space application.

Many previous works use observability indexes \cite{hollerbach1996} to optimize the choice of sampling locations. An experimental comparison of the indexes is studied in \cite{joubair2013}. An active calibration algorithm is developed in \cite{sun2008active} to find the most efficient set of calibration poses for the linearization-based techniques. A convex optimization approach to choosing configurations from a pool of candidate data points is proposed in \cite{kamali2019}. A calibration experiment design method in \cite{wu2015geometric}, based on partial pose measurement, is performed before the experiments. These methods typically involve generating a large set of randomly selected candidate configurations, followed by selecting the most optimal configurations from this set. However, actual or measured poses differ from the computed configurations due to kinematic errors, which result in robustness issues. Therefore, the calibration experimental design process should be adaptive (the choice of the next robot pose to sample should be based on previous measurements), online, and have convergence guarantees \cite{das2024bayesian}.

\section{AI SPACE CORTEX: AN INTELLIGENT FRAMEWORK FOR AUTONOMOUS SPACE EXPLORATION}
\label{sec:cortex}

\subsection{Introduction to the AI Space Cortex}
Traditional robotic space exploration systems typically rely on pre-scripted operations and deterministic state machines to execute tasks. While this may be effective for well-understood environments such as the Moon and Mars, this approach is insufficient for missions to more extreme and unknown environments, such as Europa and Enceladus, where unpredictable terrain, lesser known atmospheric properties, uncertain material properties, and long communication delays introduce high levels of operational risk.

To solve these novel challenges, we introduce the AI Space Cortex, a centralized, intelligent decision-making framework that governs all aspects of mission execution. Unlike previous approaches that largely rely on deterministic task execution, the AI Space Cortex integrates hierarchical control, active vision-based scene analysis, real-time telemetry assessment, self-diagnosing, and recalibration capabilities. This framework enables the robotic system to \textbf{autonomously initiate and execute} science missions based on real-time environmental conditions and system health assessments. Using a combination of foundational computer vision models and large language models (LLM), the Cortex can \textbf{interpret and reason} about its surroundings, making informed decisions without direct human intervention.

In addition to environmental analysis, the AI Space Cortex \textbf{detects faults} in \textbf{real-time}, \textbf{self-diagnoses} the source of anomalies, and executes \textbf{recovery} procedures as necessary. This fault tolerance ensures that the system remains operational during unexpected mechanical or sensor failures. Furthermore, the Cortex dynamically adapts its behavior based on mission constraints, environmental risks, and operational safety requirements, optimizing for both scientific value and long-term system integrity. 

\subsection{Architecture Overview}
An overview of the AI Space Cortex's system architecture and subsystems can be seen in Fig~\ref{fig:main-figure}. The AI Space Cortex serves as the central processing entity that orchestrates all robotic subsystems. A \textbf{Hierarchical Controller (HC)} functions as the task execution framework, governing nominal robotic operations and executing science mission sequences based on LLM-directed decisions. By coordinating mission-critical tasks, it ensures coordinated interaction between various system components.

The \textbf{Intelligent Scene Interaction (ISI) module} integrates computer vision and large language models (LLMs) to enable semantic understanding of the nearby surface. The ISI processes environmental data, identifies scientifically relevant features, and determines optimal sampling locations. Complementing this, the \textbf{Explanation Engine (EE)} acts as the human-machine interface, providing mission control with a visualization of the AI’s reasoning, detected faults, and operational status. Through the EE, operators can interact with the lander when necessary, ensuring that autonomous decisions remain interpretable and transparent.

To maintain system integrity, the \textbf{Fault Detection System} continuously monitors the robotic architecture, identifying and flagging anomalies in physical components such as joint health. If a discrepancy is detected, it alerts the AI Space Cortex, which determines the appropriate course of action. In the event of a kinematic fault, the \textbf{Online Recalibration System} engages to autonomously correct joint errors, ensuring the robotic arm maintains accurate movement. This recalibration approach is robust against uncertainties in joint angles, offering a more precise and reliable correction with high accuracy, particularly in complex scenarios where both rotational and translational motions must be accurately calibrated.

By continuously integrating telemetry from all system sensors, the AI Space Cortex dynamically assesses environmental risks, evaluates mission feasibility, and monitors system health before dictating and executing actions. Through this structured approach, the system ensures operational resilience and adaptability in the unpredictable conditions of planetary exploration.

\subsection{Decision-Making and Reasoning Mechanisms}
A defining feature of the AI Space Cortex is its ability to reason about the environment and mission constraints before initiating actions. This is accomplished through an integrated decision-making pipeline that utilizes both a lightweight LLM payload for telemetry-based reasoning and a multimodal vision-capable LLM for higher-level scientific analysis. These components enable the system to dynamically assess mission feasibility, prioritize scientific objectives, and trigger autonomous recovery procedures when necessary.

\textbf{Integration of Lightweight LLM Payload for Telemetry-Based Autonomous Decision Making:} The lightweight LLM payload is responsible for real-time telemetry-based decision-making, ensuring that all mission-critical conditions are met before an operation is executed. Prior to initiating any science mission, the AI Space Cortex performs a comprehensive system check, evaluating key operational parameters such as battery status, fault monitoring outputs, environmental conditions, and science objectives. Battery diagnostics assess charge levels, temperature stability, and voltage fluctuations to confirm that the system can sustain the planned mission. Simultaneously, the Cortex processes fault monitoring data, detecting potential sensor failures, joint anomalies, or hardware discrepancies that could interfere with robotic functionality. Environmental analysis further refines the decision-making process by accounting for terrain features and operational risks, ensuring that the robotic platform remains stable during excavation. Once these conditions are analyzed, the AI Space Cortex determines whether mission execution is feasible. If all parameters meet pre-trained operational thresholds, the Cortex authorizes the science mission. However, if a fault or hazard is detected, the system autonomously engages self-diagnostic and recovery protocols to resolve the issue before proceeding.

\textbf{Integration of Multimodal Large Language Models for High-Level Reasoning:} In addition to telemetry-based decision-making, the AI Space Cortex integrates a multimodal vision-capable LLM for high-level scientific reasoning, enabling it to interpret and contextualize data from its visual perception systems. Unlike traditional rule-based approaches, which rely on predefined thresholds for site selection, this LLM-based reasoning engine synthesizes segmented scene data from computer vision foundation models to generate meaningful scientific assessments. Upon receiving an image of the exploration environment, the LLM analyzes the segmented regions, extracting key geological features indicative of high scientific value. It then ranks potential exploration zones based on multiple criteria, including their estimated scoopability, material composition, and overall system safety. To ensure transparency, the AI Space Cortex provides justifications for each decision, allowing human operators to trace the reasoning process and, if necessary, override automated selections.

For instance, before finalizing a site for sample acquisition, the Cortex employs the LLM to evaluate segmented terrain images and determine the optimal collection site. The system prioritizes locations that exhibit signs of hydrated minerals, fine-grained soil, or other geological features associated with past liquid interactions. In cases where multiple sites meet mission criteria, the Cortex selects the most viable candidate by weighing scientific interest against operational constraints. Once a decision is reached, the system provides an explanation, such as: "This site is high-priority due to possible hydrated minerals, and the material appears loose enough for collection."

This multimodal reasoning capability allows planetary explorers to make context-aware, self-validated decisions rather than executing pre-scripted commands. By integrating both telemetry-based LLM logic for low-latency mission feasibility checks and vision-based reasoning for high-level scientific assessment, the AI Space Cortex ensures that mission operations are both dynamically responsive and scientifically sound.

\subsection{Autonomous Science Mission Execution}
One of the AI Space Cortex’s primary functions is to initiate and manage autonomous science missions. A science mission in the context of our system consists of a sequence of robot operations that culminate in the successful collection and storage of scientifically relevant environmental samples. The standard mission workflow follows these stages:
\begin{enumerate}
    \item \textbf{Pre-Mission Checks}
    \begin{itemize}
        \item The AI Space Cortex verifies system health and environmental conditions.
        \item Ensures available battery power is sufficient for mission completion.
        \item Ensures all fault detection and recalibration systems are active and ready.
    \end{itemize}
    \item \textbf{Scene Analysis and Target Selection}
    \begin{itemize}
        \item The mast camera captures RGB and depth images of the exploration area.
        \item The SAM-1 segmentation model \cite{kirillov2023segment} processes the images, isolating individual unique objects.
        \item The LLM-based reasoning engine evaluates the segmented objects and selects viable targets based on scientific interest and scoopability.
    \end{itemize}
    \item \textbf{Probing and Material Analysis}
    \begin{itemize}
        \item The cone penetrometer tool is utilized to measure geotechnical properties at each selected site.
        \item The AI Space Cortex filters out unsuitable sites, such as hard bedrock or potentially dangerous sites, using the force data from the cone penetration tests as a second-layer check.
    \end{itemize}
    \item \textbf{Sample Collection and Storage}
    \begin{itemize}
        \item The penetrometer is swapped for a square scoop, and the robotic arm executes a precision scoop maneuver.
        \item A visual confirmation via the mast cam ensures sample delivery to the cache.
    \end{itemize}
    \item \textbf{Mission Completion and Data Transmission}
    \begin{itemize}
        \item The AI Space Cortex logs mission telemetry and metadata, including site selection rationale, penetrometer force data, and cached sample properties.
        \item Relevant mission data is transmitted to Earth via the Explanation Engine.
    \end{itemize}
\end{enumerate}
A visual overview of this flow is given in Fig~\ref{fig:main-figure}.

\subsection{AI Personality and Adaptability in Exploration Strategies}
A unique feature of the AI Space Cortex is its ability to adapt exploration behavior dynamically. Through LLM-driven reasoning, operators can modify the mission's exploration strategy by specifying behavioral preferences, such as:
\begin{itemize}
    \item \textbf{"Conservative Mode"} – Prioritizes mission longevity, avoiding high-risk excavation sites.
    \item \textbf{"Scientific Curiosity Mode"} – Selects sites based on scientific interest, even if higher risk.
    \item \textbf{"Adventurous Mode"} – Explores aggressively, attempting riskier sample collection operations.
\end{itemize}

For example, a ground operator could instruct the AI Space Cortex to "be more aggressive in its selection of sample sites for the next 72 hours," allowing the system to dynamically adjust its behavior without requiring pre-programmed instructions for each individual mission. This capability enables real-time, autonomous adaptation based on evolving mission priorities, and optimizes site selection strategies. By incorporating this level of flexibility, the AI Space Cortex ensures that autonomous exploration is not rigidly constrained by pre-defined scripts but instead operates as a responsive system capable of modulating its decision-making framework in accordance with mission objectives and environmental variability.

\subsection{Computational Architecture}
The AI Space Cortex operates as a centralized decision-making engine within a ROS Noetic framework, interfacing with multiple subsystems via distributed ROS nodes. Each major subsystem (Hierarchical Controller, Vision System, Fault Monitor, etc.) operates as an independent network of ROS nodes, publishing and subscribing to relevant topics. Data exchange occurs through a combination of service calls (for immediate requests), publishers/subscribers (for real-time streaming data), and action servers (for long-running tasks such as arm movements facilitating science sampling missions).

The AI Space Cortex maintains an internal priority queue, where real-time telemetry updates (e.g., battery status, actuator health) are given higher priority than lower-frequency updates (e.g., environmental assessments). Scene segmentation and LLM reasoning are CPU core and GPU demanding and are scheduled based on available computational resources, ensuring that high-priority control loops (e.g., force-torque probing) remain responsive.

The AI Space Cortex is designed to operate on a system equipped with an 8-core CPU, 8GB RAM, and an NVIDIA 6GB laptop GPU. This is sufficient compute for efficient and real-time decision-making. The system’s primary control loops, responsible for actuator commands and health monitoring, operate at frequencies between 0.2 to 1.0 kHz, enabling rapid feedback and continuous stability assessments. The scene analysis pipeline, which includes SAM-1 segmentation \cite{kirillov2023segment}, runs asynchronously, updating at approximately 1 Hz when invoked, ensuring real-time perception without interfering with higher-priority control tasks. For high-level reasoning and scientific analysis, LLM queries leveraging the ChatGPT-4o API \cite{achiam2023gpt} are scheduled at a rate of 0.1 to 0.5 Hz, allowing the system to incorporate advanced decision-making without impacting the responsiveness of critical robotic subsystems.

This hardware configuration was chosen as a realistic off-the-shelf analog to flight-relevant single-board computers, such as the Qualcomm Snapdragon series of the Snapdragon’s Legacy versions (801) currently operating aboard the Mars Helicopter (Ingenuity) \cite{balaram2021ingenuity}, demonstrating the feasibility of deploying AI-driven robotic autonomy on low-power, computational hardware. By leveraging commercially available components with computational profiles comparable to those used in active planetary missions, the AI Space Cortex provides a scalable and adaptable architecture that can be optimized for future spaceflight applications.
\begin{figure*}[t] 
    \centering
    \includegraphics[width=1\linewidth]{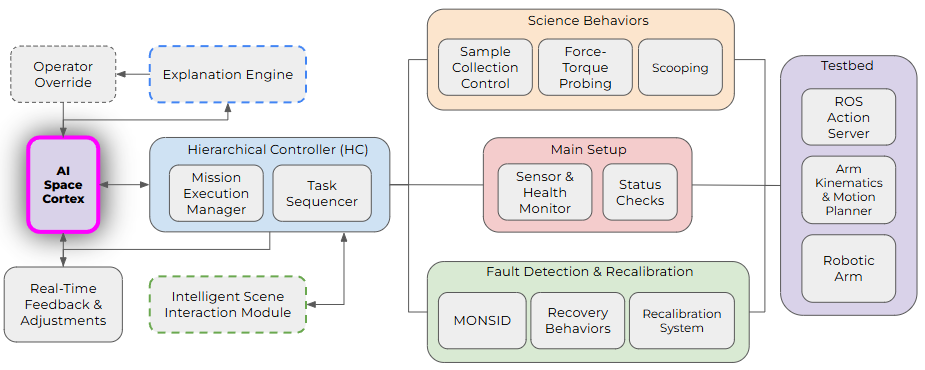}
    \caption{Hierarchical Controller (HC) flow diagram within the AI Space Cortex. The HC orchestrates mission execution, fault handling, and real-time task sequencing by integrating multiple subsystems. The AI Space Cortex issues high-level directives, which are processed by the HC and distributed across key components, including the Mission Execution Manager, Sensor and Health Monitor, and ROS Action Server. The system dynamically adapts by incorporating real-time feedback from force-torque probing, arm kinematics, and recalibration processes, ensuring safe and efficient robotic operations. The Fault Handling and Recovery module detects anomalies and engages corrective measures, such as recalibration or mission adjustments. The structured flow between components enables autonomous decision-making, fault resilience, and adaptive task execution during planetary exploration missions.}
    \label{fig:figureHC}
\end{figure*}

\subsection{Modularity for Local Model Processing}
The current implementation utilizes an API-based LLM/MLLM (GPT-4o) for high-level reasoning. Future implementations will explore fully local LLM execution, leveraging models such as Llama or emerging  Vision Foundation Models. Transitioning from an API-dependent framework to a fully autonomous, locally executed AI system presents several key challenges that must be addressed for spaceflight applications.

Running a full-scale LLM on space-relevant hardware locally requires the application of optimization techniques such as quantization to reduce model size. Additionally, there is a trade-off between inference speed and energy efficiency, as planetary landers operate under strict power constraints that limit the feasibility of high-frequency AI execution. The AI Space Cortex must balance its computational demands against available energy reserves to ensure that mission-critical functions remain unaffected. Another fundamental challenge is model adaptability, as future versions of the AI Space Cortex should be capable of dynamic updates and fine-tuning based on new mission constraints. This would eliminate the need for complete model retraining and retain the adaptability to evolving scientific objectives and environmental conditions.

\section{HIERARCHICAL CONTROLLER: THE AI SPACE CORTEX’s EXECUTION BACKBONE}
\label{sec:hierarchical}

\subsection{Introduction to the Hierarchical Controller}
The Hierarchical Controller (HC) serves as the execution framework for the AI Space Cortex, providing low-level robotic control and task sequencing. While the AI Space Cortex makes high-level mission decisions, the HC translates these decisions into concrete robotic actions, ensuring that operations are safely and efficiently executed. If desired, the HC can be driven by predefined scripts, though AI reasoning provides more flexible operations. A visual overview of its architecture can be seen in Fig~\ref{fig:figureHC}.

\subsection{Role of the Hierarchical Controller in the AI Space Cortex}
The Hierarchical Controller (HC) follows the AI Space Cortex’s decision-making logic. It also coordinates system health checks and monitors sensor feedback, and maintains situational awareness of the robotic platform’s operational status. Additionally, it handles real-time state transitions within a customized ROS Noetic framework, allowing for seamless execution of sequential and parallel robotic processes.

To facilitate scalable task management, the HC distributes execution responsibilities across multiple robotic subsystems using ROS action servers, ensuring that each component subsystem performs its designated function without conflict. Throughout mission execution, it continuously provides real-time feedback to the AI Space Cortex, enabling adaptive mission planning based on system performance and environmental constraints. While the HC plays a vital role in managing task execution, it does not independently make mission-critical decisions. Instead, it functions as an intelligent execution layer, ensuring that AI-directed commands are carried out with precision, safety, and system-wide coordination.

\textbf{System Architecture and ROS Noetic Implementation:} The Hierarchical Controller is built within the ROS Noetic framework, with a modular architecture that facilitates scalability and adaptability.

\textbf{High-level ROS Node Structure:} The HC operates as a collection of ROS nodes that manage different aspects of robotic control. These nodes operate state transitions asynchronously, ensuring that different subsystems can execute tasks independently.

\textbf{Communication with the AI Space Cortex:}
The HC communicates with the AI Space Cortex through a combination of ROS topics, service calls, and action servers. This structure ensures efficient execution while maintaining real-time feedback loops to prevent errors or unsafe actions.

\subsection{Execution of a Science Mission}
A Science Mission consists of a structured sequence of tasks, each executed under the control of the HC while being supervised by the AI Space Cortex. The workflow follows a hierarchical state transition model, where each stage is monitored for completion, failure, or unexpected anomalies.

\textbf{Science Mission Workflow:} The overall architecture of a typical mission can be observed in Fig~\ref{fig:figureHC}, and an overall, end-to-end overview of an entire mission can be observed in Fig~\ref{fig:main-figure}. This structured, hierarchical execution model ensures that each operation is completed safely, with continuous monitoring and dynamic adaptation.

\subsection{Fault Detection and Recovery in the Hierarchical Controller:} The Hierarchical Controller (HC) implements real-time fault detection mechanisms to prevent execution errors and ensure system stability during mission-critical operations. By continuously monitoring hardware integrity and environmental constraints, the HC detects, isolates, and responds to anomalies before they escalate into mission failures.

\textbf{Fault Detection Strategies:} To maintain operational reliability, the HC monitors manipulator arm joint health by tracking encoder feedback, joint torques, and overheating risks, ensuring that any abnormal deviations are identified before they impact mission execution. Battery protection mechanisms regulate power distribution, preventing high-consumption actions from being executed when energy reserves fall below safe thresholds. Additionally, sensor redundancy checks safeguard against primary sensor failures; if a primary sensor becomes unresponsive, the system either attempts a reset or switches to a backup sensor, such as a secondary RGB-D camera, to maintain situational awareness. In the event of unexpected environmental obstructions, dynamic execution preemption allows the HC to immediately pause all motion and request intervention from the AI Space Cortex, ensuring that hazardous interactions are avoided.

\textbf{Recovery Protocols:} Once a fault is detected, the HC follows a structured recovery protocol based on fault severity and system constraints. For minor faults, such as transient joint errors or temporary signal loss, the HC automatically retries the operation, attempting multiple execution cycles before escalating the issue. Medium-severity faults, such as power fluctuations or unexpected sensor discrepancies, cause the HC to pause execution and await AI Space Cortex intervention to determine whether the mission should continue, be modified, or enter a lower power state. If the system encounters a critical but fixable fault, such as a misaligned joint that can be recalibrated, the HC halts all movement and signals the AI Space Cortex to initiate a repair procedure before resuming operations. In cases of severe or unfixable faults, such as a catastrophic joint failure beyond autonomous repair capabilities, the HC enters safe mode, suspends all robotic motion, and sends an emergency request for human intervention.

These mechanisms ensure operational robustness and minimize mission disruptions caused by minor or fixable faults. The system is also able to properly escalate critical failures when necessary. By balancing autonomous fault handling and AI-guided decision-making, the HC enhances system resilience and maximizes the success rate of the planetary exploration missions.

\subsection{Adaptive Behavior and Context-Aware Decision Making}
Although the Hierarchical Controller (HC) primarily functions as an execution framework, it also incorporates context-aware adaptability, allowing it to dynamically adjust mission plans based on real-time system conditions. By continuously evaluating operational constraints, the HC ensures that science operations prioritize both safety and efficiency, maximizing mission success rates. When unexpected environmental factors or execution challenges arise, the HC can re-plan task sequences in real time, ensuring that robotic operations remain both feasible and optimized for the prevailing conditions.

For instance, if the AI Space Cortex selects a site for sampling but the force-torque probing data reveals that the surface is too hard for scooping, the HC can autonomously select an alternative site from the list of pre-validated options identified by the AI Space Cortex. This decision is made without requiring explicit intervention from the Cortex, provided the newly selected site still aligns with mission constraints. By incorporating adaptive flexibility into its execution model, the HC improves system efficiency and mission throughput, ensuring that planetary exploration remains both autonomous and resilient to environmental variability.

\section{INTELLIGENT SCENE INTERACTION: VISION-BASED PERCEPTION AND SEMANTIC UNDERSTANDING}
\label{sec:segmentation}

\subsection{Introduction to Intelligent Scene Interaction}
A fundamental requirement for autonomous space exploration is the ability to perceive, analyze, and interpret the environment in real time. Unlike Earth-based robotic systems, which benefit from human oversight, predefined maps, and high-powered compute clusters, planetary exploration robots must operate in unknown, unstructured environments with minimal external input. They must also adhere to strict power and processing limitations. The ability to autonomously detect, segment, and reason about objects within the scene is therefore critical to ensuring mission success.

The Intelligent Scene Interaction (ISI) module within the AI Space Cortex provides this capability by integrating advanced vision-based segmentation, large language model (LLM)-driven semantic reasoning, and real-time material analysis through onboard sensor fusion. Computer vision models process and segment the environment, extracting meaningful features (based on pre-trained data) from real-time RGB-D data. Using LLM-based reasoning, the ISI module evaluates and prioritizes potential sampling sites by assessing their scientific relevance and operational feasibility, where relevance can be specified by a prompt drawing from pretrained models. The module also performs real-time geometric and material property analysis, allowing the system to adapt to safety variable terrain conditions.

By incorporating these capabilities, the ISI module enables the AI Space Cortex to autonomously identify and rank high-value scientific targets, evaluate terrain stability and scoopability, filter out non-viable sites using high-speed computational heuristics, and generate structured justifications for site selection based on pretrained, scientific reasoning. Through this approach, the ISI module serves as the bridge between robotic perception and AI-driven decision-making. It allows for intelligent site selection and context-aware science missions.

\begin{figure*}[t] 
    \centering
    \includegraphics[width=1\linewidth]{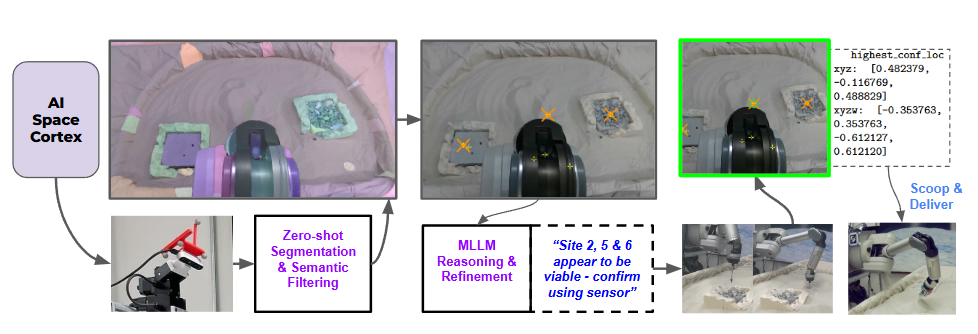}
    \caption{Intelligent Scene Interaction (ISI) module's perception and decision-making pipeline for autonomous sample site selection. The process begins with the AI Space Cortex capturing an RGB-D image via the mast camera. The raw image is processed through a zero-shot segmentation model, producing a fully segmented representation of the test site. Semantic filtering techniques are applied to remove non-viable objects, resulting in a refined selection of potential sample locations. The AI Space Cortex then employs a multimodal LLM for high-level reasoning, identifying scientifically promising sites (highlighted with orange crosses) and recommending sensor-based validation. The robotic system executes force-torque probing at the selected locations, confirming material properties before finalizing the sampling site (marked with a green circle). Cartesian coordinates of the confirmed site are displayed, aligning with the system’s kinematic reference frame to ensure precise robotic interaction. The entire pipeline—from image capture to final site confirmation—was completed in approximately 8 minutes and 40 seconds.}
    \label{fig:figureA2}
\end{figure*}

An end-to-end visualization of the ISI module's full capability is shown in Fig~\ref{fig:figureA2}.

\subsection{System Architecture}
The Intelligent Scene Interaction (ISI) module operates within a ROS Noetic framework and consists of three primary subsystems, each designed to contribute to the end-to-end perception, reasoning, and decision-making pipeline for autonomous science operations.

The computer vision pipeline extracts RGB-D data from onboard cameras and applies foundational model-based image segmentation to identify and classify distinct objects within the scene. The system combines depth data and color imagery to ensure accurate spatial localization of potential sampling sites.

Once initial segmentation is completed, the Semantic Filtering and Scene Refinement subsystem processes the segmented scene, utilizing geometric constraints, image engineering, and heuristic-based filtering techniques to eliminate non-viable objects. This step ensures that the ISI module prioritizes scientifically relevant and physically accessible targets and removes outliers that would either pose a mechanical risk or hold little exploratory value.

Finally, the AI-Driven Scene Analysis subsystem integrates LLM-based reasoning to evaluate the scientific significance and overall operational feasibility of selected sites.

\textbf{ROS Node Structure:} The ISI module is structured as a collection of ROS nodes, allowing for parallelized processing. The most prominent include:
\begin{itemize}
    \item \texttt{/vision\_processing} – Captures RGB-D images and processes them for segmentation.
    \item \texttt{/scene\_segmentation} – Runs the SAM-1 segmentation model to isolate objects in the environment.
    \item \texttt{/object\_filtering} – Applies heuristics and geometric constraints to eliminate non-viable objects.
    \item \texttt{/llm\_reasoning} – Calls the GPT-4o API, retrieving real-time site rankings, reasoning justifications, and uncertainty metrics for exploration targets.
    \item \texttt{/probe\_analysis} – Handles force-torque probing to verify material properties.
\end{itemize}
This modular structure enables real-time interaction, with data streams updating asynchronously at different frequencies based on computational constraints.

\subsection{Computer Vision Pipeline}
\textbf{RGB-D Image Acquisition:} The Intelligent Scene Interaction (ISI) module relies on an RGB-D mast camera to capture a multi-perspective view of the terrain, providing both color imagery and depth data to enhance scene understanding. The depth data is essential for precise object localization and terrain reconstruction, enabling the robotic arm to accurately assess target sites for interaction. The image acquisition process follows a structured sequence, beginning with camera calibration and gimbal adjustments to ensure that the sensor alignment is optimized for scene capture. Once calibration is complete, the system performs RGB-D image acquisition from a single, controlled viewing angle, capturing a high-resolution snapshot of the exploration environment. The final step in this pipeline involves processing layer alignment, ensuring that the image and depth data are precisely synchronized with the robotic system’s kinematic pipeline, allowing for accurate real-world localization of detected objects.

\textbf{Segmentation Using the SAM-1 Vision Transformer:} After image acquisition, the captured RGB-D data is processed through the SAM-1 vision transformer-based segmentation model, developed by Meta AI \cite{kirillov2023segment}. This zero-shot generalization model autonomously identifies and isolates objects within the scene without requiring additional training, making it highly suitable for exploration in previously unknown planetary environments. The segmentation model outputs a structured representation in which each detected object is treated as a unique entity, forming a structured scene graph that serves as the foundation for semantic filtering and scientific evaluation. By leveraging the capabilities of modern vision transformers, this pipeline ensures that the AI Space Cortex can efficiently analyze complex planetary terrains.

\subsection{Semantic Filtering and Scene Refinement}

\textbf{High-Speed Computational Filters:} The SAM-1 segmentation model can generate hundreds of segmented objects in a single scene, necessitating real-time filtering techniques to eliminate non-viable candidates and improve computational efficiency. To achieve this, the ISI module applies a series of filters designed to optimize the selection of scientifically and operationally viable targets.

The size constraint filter removes objects that are either too large or too small for robotic interaction, based on an understanding of the end-effector's geometry, which we hard-coded into the filter, ensuring that only realistically scoopable objects are considered. The spatial proximity constraint filter eliminates objects that are too close to one another, based again on a pre-loaded understanding of the end-effector geometry, reducing the risk of unintended collisions or sampling interference. To prevent false-positive site selections, the material classification filter excludes objects identified as metallic, artificial debris, or robotic system components, which are non-scientific artifacts. Finally, the reachability constraint filter discards any objects located outside the robotic arm’s effective operational range by computing the max-extension polygon, ensuring that selected sites are physically accessible for sampling. By applying these real-time constraints, the ISI module reduces the number of viable targets from hundreds to a manageable subset of a few dozen or less, optimizing both decision-making efficiency and mission execution speed.

\textbf{Large Language Model (LLM) Integration:} Once the computational filtering stage is complete, the AI Space Cortex employs LLM-based reasoning to refine its selection and rank the most promising sites. This process begins by transmitting refined object metadata, including size, location, depth, and material composition estimates, to the LLM-based scientific reasoning module. The LLM (multimodal GPT 4o from OpenAI in this case) - then evaluates each candidate site by comparing observed features to pre-trained geological and environmental datasets, determining which sites exhibit the highest likelihood of scientific significance. Scientific significance is inferred based on the model's pre-trained data, and payload prompts that the operator is encouraged to pre-load, e.g. "Hey! We don't want you to risk our scoop today. The mission is still young, be careful, we only want to collect a small amount of something soft." The implication being that the model will understand to prioritize softer looking sediment, avoid objects with complex curvatures, avoid large rocks, etc. Based on this analysis, the Cortex assigns a priority ranking to each object, incorporating perceived significance of locations, mission objectives, and operational constraints into its assessment.

As part of this process, the LLM generates a natural language justification for the selected sites, allowing for full traceability of AI-driven decision-making. While the 4o, OpenAI model is a multimodal model and is pretrained on images and text, it isn't performing a strict mineralogical classification based on physical principles. Rather, it is inferring \textit{likely} scientific value based on information that it learned during its extensive training. For example, the ISI module may output:
\begin{quotation}
"The segmented object at site-7 is a potential high-value target due to its wavy distribution, which suggests a possible soft deposition. The material appears to be fine-grained, like quartz, making it a suitable candidate for sample collection. Confidence level: 8/10." - \textit{AI Space Cortex}
\end{quotation}
By combining machine vision with LLM-based scientific reasoning, the AI Space Cortex allows the system to contextualize its observations and make informed, mission-driven decisions rather than executing purely predefined heuristics. This ensures that selected targets are scientifically relevant and operationally feasible, and advances the current level of autonomous planetary exploration capabilities.

\subsection{AI-Driven Scene Analysis and Exploration Decision Making}
\textbf{Probing and Material Verification:} Before committing to a sampling action, the AI Space Cortex employs a force-torque penetrometer to validate the physical properties of potential sampling sites. The system initiates this process by commanding the robotic arm to press the penetrometer into each candidate site, measuring depth resistance and material hardness to assess scoopability. By analyzing force feedback data, the system determines whether the terrain is composed of loose, granular material or obstructed by a compacted or bedrock layer, ensuring that only viable locations are considered for sample collection. Based on this assessment, the AI Space Cortex generates a final boolean estimate, confirming whether a site meets mission constraints for excavation. This decision-making process prevents unnecessary execution attempts at sites where retrieval would be mechanically infeasible or possibly lead to equipment damage.

\textbf{Final Site Selection and Scoop Execution:}
Once material verification is complete, the AI Space Cortex finalizes its site selection by ranking candidate locations based on scientific value, scoopability, and mission constraints. The system prioritizes sites that exhibit high scientific potential, such as those indicative of past hydrological activity. The system also ensures that the selected location is physically accessible and operationally safe for robotic interaction. Additional factors such as battery levels, available mission time, and overall sampling priority are also considered before proceeding with excavation.

With the final target identified, the robotic arm executes the scoop maneuver to collect a sample from the designated location. The collected material is transported to the storage cache, ensuring that it remains preserved for further analysis. Throughout this process, the Intelligent Scene Interaction (ISI) module continuously logs metadata, including RGB-D image snapshots, depth measurements, and force feedback data, providing a complete dataset for post-mission review. This information is also integrated into the Explanation Engine, allowing human operators to visualize mission execution and validate AI-driven decisions.

\section{EXPLANATION ENGINE}
\label{sec:explanation}

\subsection{Introduction to the Explanation Engine}
The Explanation Engine (EE) serves as the human-machine interface for the AI Space Cortex, ensuring that ground operators can comprehend, validate, and interact with the system’s autonomous decision-making processes. Unlike traditional planetary exploration missions, where human operators are primarily responsible for directing high-level robotic actions, the AI Space Cortex functions as a fully autonomous decision-maker. However, complete autonomy without explainability introduces significant risks, as the inability to interpret why the system makes certain choices can erode trust in AI-driven decision-making and hinder operational oversight.

The Explanation Engine addresses this challenge by providing a structured interface that enables mission control to observe, analyze, and intervene when necessary. The EE offers real-time visualization of system states, mission progress, and AI Space Cortex decision logic. This ensures that operators can track the system’s reasoning at every stage. Through graphical overlays, the EE illustrates site selection rationales, uncertainty assessments, and detected faults, allowing for rapid situational awareness. In addition to decision tracking, the EE streams live telemetry, including joint status, battery levels, and power distribution metrics, enabling ground operators to monitor critical system health parameters. If required, the EE also facilitates interactive overrides, allowing mission control to adjust mission parameters or modify AI-selected targets when deemed necessary.

\begin{figure*}[t] 
    \centering
    \includegraphics[width=1\linewidth]{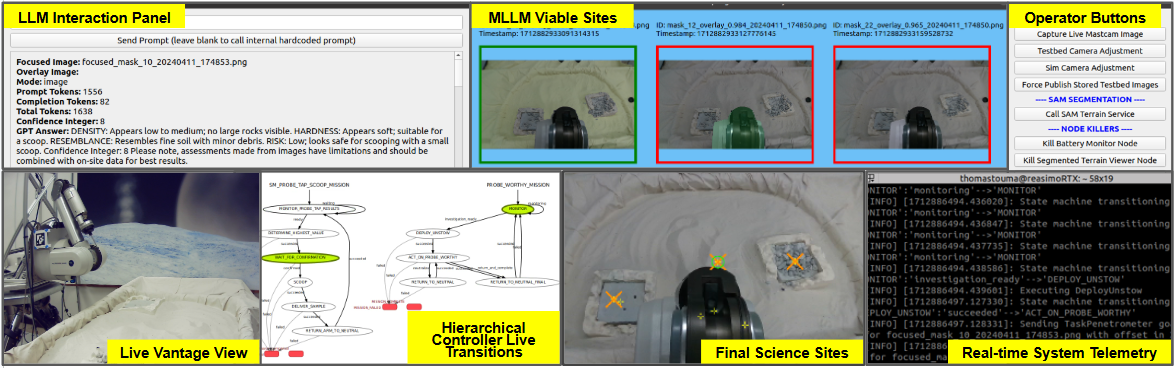}
    \caption{Graphical user interface of the Explanation Engine (EE) amalgamated neatly, providing real-time visualization and operator interaction tools for AI Space Cortex driven exploration. The LLM interaction panel enables dynamic AI personality adjustments and allows operators to issue direct queries regarding mission reasoning and site selection confidence. The MLLM viability panel displays identified sampling sites, ranked by scientific viability, using green (high confidence), orange (moderate confidence), and red (non-viable) markers. An operator control panel provides manual override options for functions such as image capture, mast adjustment, and robotic arm intervention. A live vantage camera feed offers a secondary perspective for safety monitoring during operations. The ROS Noetic SMACH transition block diagram visualizes live state transitions within the Hierarchical Controller, allowing real-time tracking of task execution. The final sampling sites are displayed with orange crosses for candidate locations and green circles denoting confirmed and force-probed scoop targets, ensuring transparency in the AI Space Cortex's decision-making. Additionally, a live telemetry output terminal streams system diagnostics, power status, and execution logs, offering operators a detailed view of ongoing mission activities.}
    \label{fig:figureEE}
\end{figure*}

By integrating advanced visualization tools, structured LLM-generated output, and interactive user-control mechanisms, the EE ensures that the AI-driven exploration process remains transparent and traceable to the operator. This enhances mission reliability and strikes a balance between human oversight and AI-driven decision-making in our autonomous space exploration systems. An overview of the Explanation Engine's graphical user interface (GUI) is shown in Fig~\ref{fig:figureEE}.

\subsection{System Architecture and ROS Integration}
The Explanation Engine is built on a ROS Noetic-based architecture, interfacing with multiple subsystems to aggregate, format, and visualize data for human operators.

\textbf{ROS Node Structure:} The EE consists of several primary ROS nodes. The /ee\_interface node serves as the central processing node, responsible for data aggregation and GUI updates. The  /llm\_ee\_handler retrieves formatted justifications from GPT-4o/GPT-4-preview API outputs, structuring them into readable formats. The /scene\_visualizer generates real-time overlays of site selection rankings and confidence metrics. The /fault\_ee node displays joint health, sensor errors, subsystem anomalies, and MONSID data. The/telemetry\_terminal node streams live mission status, including battery, temperature, and signal integrity data. Each node subscribes to relevant topics from the AI Space Cortex and updates the graphical user interface (GUI) asynchronously.

\textbf{Communication with the AI Space Cortex:} The Explanation Engine (EE) retrieves and processes data from various AI Space Cortex subsystems, ensuring that mission control has real-time access to telemetry, justifications, and interactive control mechanisms.

For continuous telemetry streaming, the EE subscribes to ROS Topics, receiving high-frequency updates ranging from 1 Hz to 100 Hz, depending on the system parameter being monitored. This includes joint health data, battery status, force-torque measurements, and system alerts, allowing operators to maintain full situational awareness of the robotic platform. In addition to real-time monitoring, the EE supports on-demand queries through ROS Services, enabling it to request justifications from the LLM module when operators need further explanation regarding site selection, mission prioritization, or system responses to detected anomalies.

Beyond passive monitoring, the EE also provides interactive override mechanisms using ROS Actions, granting operators the ability to pause, modify, or redirect robotic tasks as needed. These interactive controls allow mission control to issue commands for stopping or repositioning the robotic arm, as well as receive goal movement feedback. This ensures that operator interventions are executed precisely and that they receive structured, timely, and actionable feedback, maintaining a balance between AI autonomy and human oversight in space missions.

\subsection{Explanation Framework for AI Space Cortex Decisions}
A critical function of the Explanation Engine (EE) is to ensure that AI-driven decisions are presented as traceable, and actionable for human operators. Since the AI Space Cortex leverages GPT-4o for semantic reasoning, the EE is responsible for translating raw LLM outputs into structured, meaningful insights that provide scientific justification for mission choices. Without this structured framework, AI-generated reasoning would remain verbose and difficult to analyze efficiently in real-time operational scenarios.

\textbf{Structuring LLM Justifications for Operators:} Rather than displaying unprocessed text responses from the LLM, the EE formats justifications into structured outputs, presenting key decision factors in a concise yet scientifically rigorous format. The structured output includes metrics for the likelihood of success, such as scoop success probabilities represented as normalized scores (e.g., "..Confidence Level: 9/10" [Scoop success probability]), providing operators with an immediate quantitative assessment of site viability. Additionally, the EE displays computed reasoning and context on why a site was chosen based on pre-trained, inferred, statistical association of planetary science knowledge (e.g., "[perceived] Layered formations suggest subsurface ice deposits"). The statistical scientific prioritization ranking further refines the selection by categorizing sites in order of perceived importance (e.g., "Ranked \#1 due to possible hydrated minerals"), allowing mission control to quickly validate AI-driven decisions. Based on its training and natural language reasoning capabilities, the LLM model from OpenAI will rank sediment-like features higher than large rocks, particularly when guided by operator prompts that encourage it to be conservative in that mission. Finally, the EE displays assigned uncertainty scores to indicate confidence in the LLM's assessment (e.g., "Low confidence due to partial occlusion in imaging"), ensuring that ambiguous decisions are flagged for further verification.

\textbf{Decision Explanation Flow:} To facilitate this structured decision-making process, the EE follows a sequential explanation flow when handling AI Space Cortex queries. The process begins when the AI Cortex submits a request to the /llm\_ee\_handler, querying GPT-4o for a justification regarding site selection. Upon receiving the raw response, the EE parses and structures the output, assigning confidence scores derived from pre-trained datasets and historical mission data. The formatted explanation is then displayed in the EE graphical user interface (GUI), presenting operators with a clear, structured rationale for the AI’s decision. If the assigned uncertainty score exceeds a predefined threshold, the EE automatically flags the site for additional verification, prompting the AI Cortex to initiate further data collection, such as re-running force-torque probing or requesting an alternative image capture for refinement.

By implementing this structured explanation framework, the EE ensures that human operators can quickly interpret AI-driven decisions, validate scientific reasoning, and intervene when necessary.

\subsection{Real-Time Visualizations and Operator Dashboard}
\label{subsec:dashboard}
The Explanation Engine (EE) serves as more than just a textual interface—it provides a fully interactive graphical user interface (GUI) that visually represents mission data, AI decision rationales, and system health monitoring in real time. By incorporating structured overlays and operator-interactive controls, the EE ensures that ground operators can effectively interpret and validate AI-driven decisions as they occur. Override capabilities, described in Section~\ref{subsec:dashboard} allow operators to intervene when manual control is desired.

\textbf{Scene Visualizer (Site Selection Overlay):} To facilitate real-time analysis of sampling site selections, the EE displays an interactive site selection overlay, allowing operators to visualize the AI Space Cortex’s decision-making process. High-priority sampling sites, which exhibit an 80\% or higher likelihood of scoop success, are highlighted in green, indicating areas that the system has deemed optimal based on force-torque probing and LLM assessments. Medium-confidence sites, with success probabilities between 60-79\%, appear in orange, denoting areas that require additional sensor validation before final selection. Sites classified as non-viable, such as regions where bedrock has been detected or where scoopability is deemed infeasible, are marked in red, ensuring that mission control can quickly discern which regions are not suitable for excavation.

\textbf{Fault and System Health Monitoring:} Beyond site visualization, the EE continuously receives system health parameter data, ensuring real-time feedback on the robot's status. Any joint or actuator failures are immediately flagged and displayed in live fault overlays, allowing operators to diagnose mission hazards before they compromise execution. Critical power-related issues, such as low battery warnings, are visualized within the GUI terminal and are automatically logged in the panel’s telemetry output. Additionally, the EE provides telemetry alerts for a range of system integrity factors, including thermal management anomalies, bandwidth degradation in local testbed networks, and unexpected motion feedback, ensuring that any deviations from nominal operating conditions are detected and addressed in real-time, if needed.

\textbf{Operator Override Panel:} Although the AI Space Cortex is designed for autonomous decision-making, the EE includes an operator override panel, allowing mission control to intervene when necessary. Through this panel, operators can override AI site selection, manually selecting an alternative site if mission requirements dictate a deviation from the Cortex’s automated decision. Additionally, operators can adjust risk thresholds dynamically, permitting scooping operations in medium-confidence zones if mission constraints—such as time or remaining power—require a more aggressive approach to sample collection. Lastly, the override panel allows operators to re-prioritize mission goals, switching between Adventurous Mode, Scientific Curiosity, and Conservative Mode if telemetry data indicates that power reserves are depleting too rapidly to sustain high-risk operations.

\subsection{Explanation Engine Computational Performance}
Since GPT-4o queries introduce inherent latency constraints, the Explanation Engine (EE) implements multiple performance optimizations to ensure that AI-driven explanations are delivered efficiently and in near real-time, without overloading system resources. Batch processing significantly reduces computational overhead by grouping multiple queries into a single request rather than processing each site selection individually. This approach optimizes response time and minimizes redundant calls to the LLM, allowing for faster decision-making during mission execution.

Additionally, the EE ensures that all LLM-generated responses are concise yet informative, thus eliminating unnecessary verbosity. For example, it provides the full explanation and also a short summary of confidence information, which is useful in the event that the operator is required to make a quick judgment. The structured output is formatted for rapid operator interpretation but also keeps all important information visible in the response. These optimizations collectively enable the EE to deliver fast AI-generated justifications with low computational burden, and ensure that mission control receives actionable insights.

\section{MONSID} 
\label{sec:monsid}
The Model-based Off-Nominal State Identification and Detection (MONSID) \cite{kolcio2016model} system was used to detect and diagnose faults in the arm actuators and sensors. A core MONSID system implementation consists of a generic diagnosis engine and a model specific to the physical system being monitored. A suite of sensor and command data from the physical system is used to feed a representation of sensors and commands in the model. The diagnosis engine includes fault detection parameters that are tuned for a particular system. For this project, we built a MONSID model of the OWLAT robot arm. This includes the arm’s joint configuration and kinematics, the joint actuators, the sensors that report the joint positions, the commands that direct joint movement, and the arm position as determined from a camera’s view of the arm. The model will be described first, followed by the MONSID ROS implementation that was utilized with the OWLAT simulation and hardware testbed, later discussed in Section~\ref{sec:owlat}. 

\subsection{MONSID's OWLAT Arm Model}
\begin{figure}[t]
    \centering
\includegraphics[width=1\linewidth]{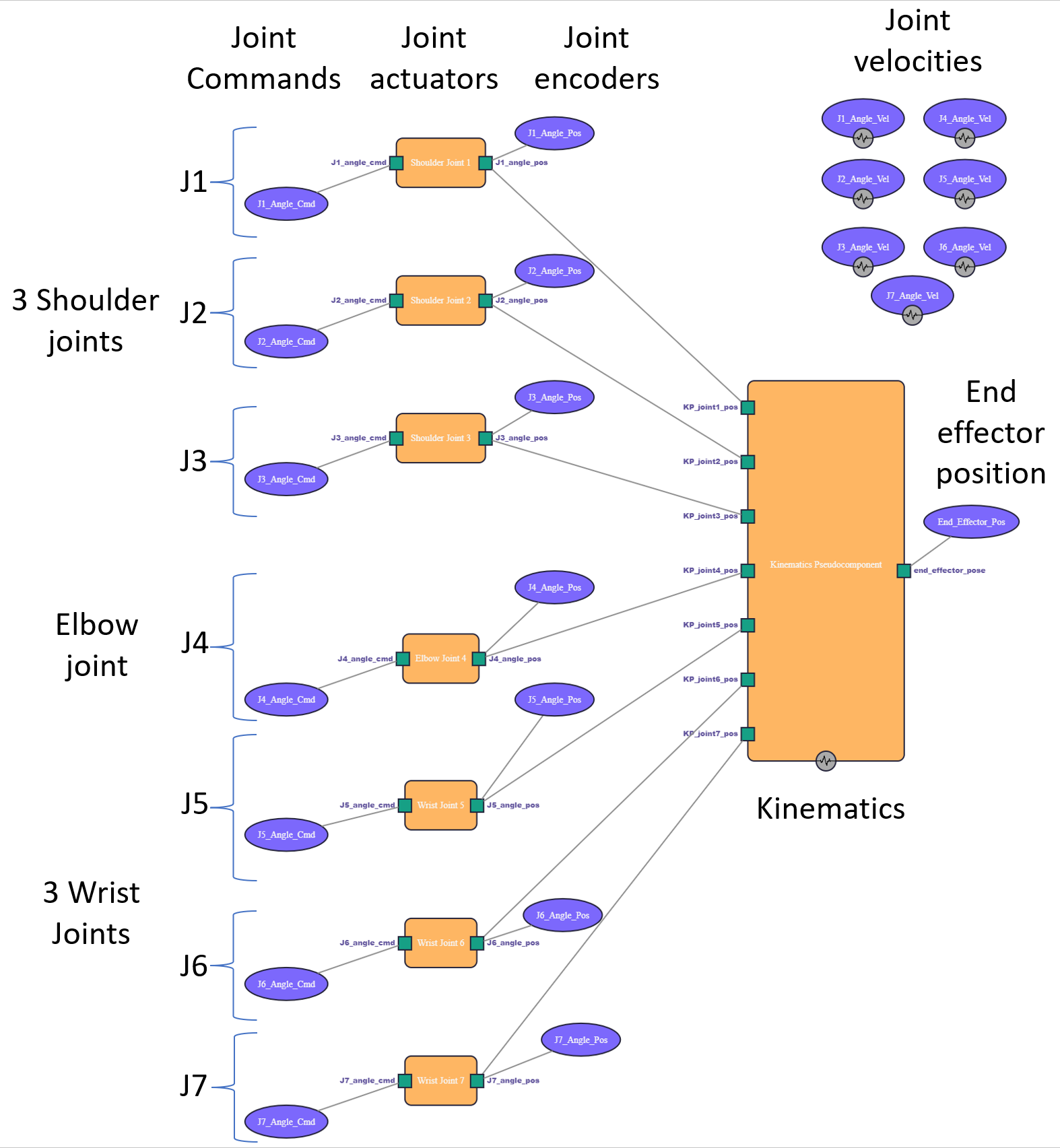}
    \caption{ MONSID model of OWLAT's arm.
    }
   \label{fig:monsid1}
\end{figure}

A MONSID model consists of interconnected components representing hardware on the physical system. Each component contains input and output ports called nodes that send and receive data that is processed by their parent component and sent out to neighboring components. Commands to and measurements from the physical hardware are inputs to the model and are sent to appropriate nodes. A topological diagram of MONSID’s model of the OWLAT arm is shown in Fig.~\ref{fig:monsid1}. The purple ovals, from left to right, are the inputs to this model: the commanded joint angular positions (determined by the OWLAT arm controller) for each of the seven joint actuators; the positions of each of joints as reported by the motor encoders; the joint angular velocities, and the position of the arm’s end-effector (as derived from the camera image of the arm). The orange boxes on the left of the figure represent the joint actuators. The joint actuator components calculate a joint angle given the commanded angle. The actuator components are connected to the large orange box on the right of the figure, which calculates the Cartesian position of the end of the robot arm (end-effector) via forward kinematics, given the joint angles. 

Each component models the expected, correct behavior of the hardware. Above, we described the components’ input-to-output behavior, known as the forward path. The components also calculate the output-to-input behavior, known as the reverse path. The reverse relationships are typically the inverses of the forward relationships. The inverse relationships in the individual joint actuator components are straightforward. The reverse relationship in the kinematics component requires calculating the joint angles. Calculating the joint angles from the end-effector’s position requires a solution to the inverse kinematics, which is not unique for a 7-joint arm. The joint velocities were integrated to obtain the joint angles in the reverse path. 
\begin{figure}[t]
    \centering
\includegraphics[width=1\linewidth]{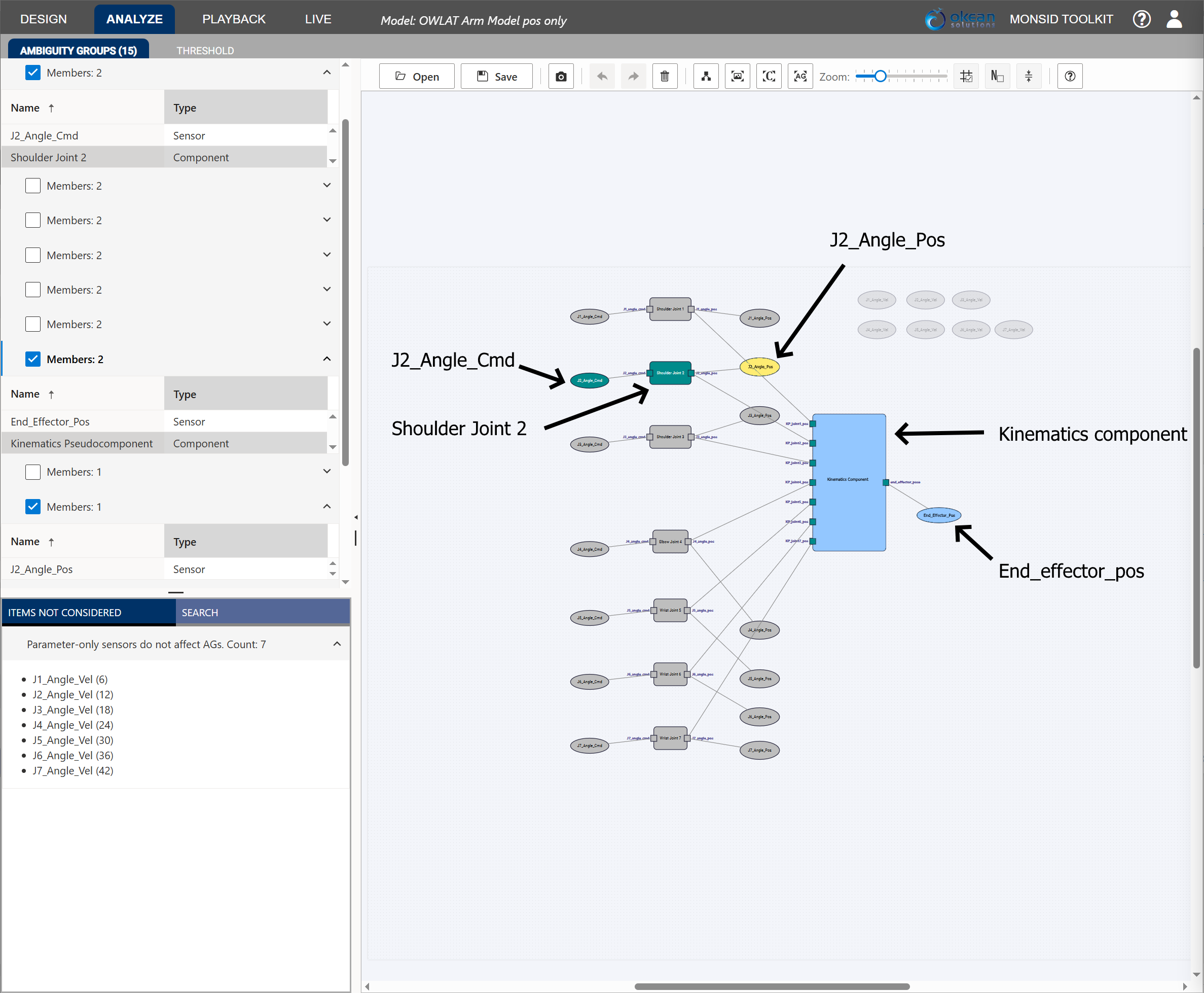}
    \caption{ Snapshot of the MONSID Toolkit ambiguity group analysis of the OWLAT model (annotated). 
Green shading shows an ambiguity group containing J2\_Angle\_Cmd and Shoulder Joint 2. 
Yellow shading shows an ambiguity group with a single member, J2\_Angle\_Pos. A third ambiguity group is shown in blue shading, consisting of the kinematics component and end-effector position sensor.
    }
   \label{fig:monsid2}
\end{figure}
\begin{figure}[t]
    \centering
\includegraphics[width=1\linewidth]{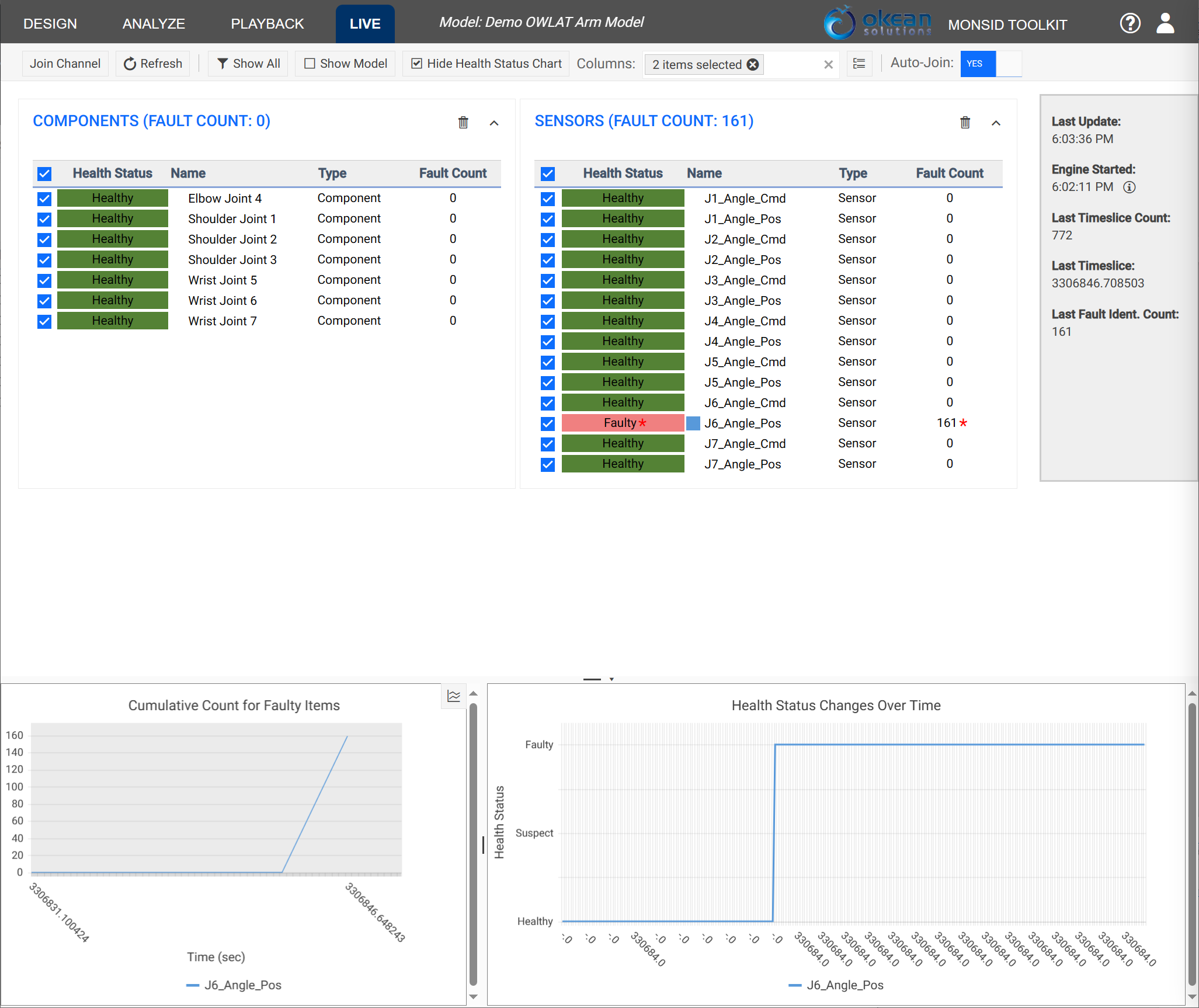}
    \caption{ Snapshot of MONSID Toolkit Livestream page showing health status during a bias fault injected into the joint 6 encoder on the OWLAT hardware arm. MONSID determined “J6\_Angle\_Pos” – joint 6’s position sensor, to be at fault.
    }
   \label{fig:monsid3}
\end{figure}

\subsection{Fault Detection and Identification}
During operations, input data is propagated in forward and then reverse paths through the model resulting in multiple values representing the same physical quantity at each node. This is done to generate analytic redundancy that is used for both fault detection and identification. At each time slice, MONSID checks the consistency of the values in each node. Health states remain healthy if all values pass consistency checks. Problems in the arm will manifest as inconsistencies. The inconsistencies potentially lead to detected faults. 

The engine is given fault detection parameters, including tolerances on the inconsistencies and their persistence, set depending on sensor noise and data rates. On a flight mission the parameters are fine-tuned based on tradeoffs between desired false positive versus mis-detection rates. On REASIMO, tolerances are determined by reviewing the noise levels of nominal data (both the sensor noise and data rate stability) collected from the hardware testbed. Only once the inconsistencies exceed the detection parameter limits, does the engine declare that a fault has been detected. In the time slice that a fault is declared, the MONSID engine also performs fault identification to determine where the fault occurred and then sets the health states of the faulty component/sensor accordingly. This technique can identify the source of faults and to distinguish between faults in hardware from faults in the sensors measuring their output. MONSID’s ability to identify the source of faults depends on the layout, or topology, of the underlying model. MONSID models can be analyzed to determine groups of model elements (components and sensors) called ambiguity groups \cite{kolcio2017model}. 

MONSID can isolate a fault to a particular ambiguity group. Thus, models whose topology results in many small ambiguity groups will have better diagnostic resolution than models with few large ambiguity groups. For a perfectly diagnosable model, each component and sensor would be the only member in its own ambiguity group. Ambiguity groups with more than a single member can be reduced with the addition of sensors, which may be possible when fault management design is closely coupled with flight system design.

On REASIMO, we are working with an as-built system with a fixed sensor suite. The completed topology of the OWLAT model was analyzed in the MONSID Toolkit (part of the MONSID tool suite) for ambiguity groups. With this topology, it should be possible to distinguish between a fault in any actuator component and the encoder reporting the actuator’s joint angle. This was verified in our experiments on the testbed. Each joint command and actuator pair form an ambiguity group, while each encoder is a single member of its own ambiguity group. The kinematics component and end-effector sensor are in their own two-member group. A screenshot of the Toolkit ambiguity group analysis page is shown in Fig.~\ref{fig:monsid2}. A few groups have been expanded in the list on the right pane and indicated by the shaded elements in the model diagram. The ambiguity group shaded green includes Shoulder Joint 2 actuator and joint angle command (J2\_Angle\_Cmd). The yellow shade indicates the joint 2 encoder (J2\_Angle\_Pos). The blue shade indicates the kinematics component and end-effector. It is also possible to distinguish between joint actuator/encoder faults and a physical fault like a bent link somewhere in the robot arm. In that case, MONSID would determine the kinematics component and end-effector to be faulty. If the camera is believed to be operating correctly, this information could then be utilized to examine images from the mast camera to help identify which arm link was bent and then adjust activities accordingly.

\subsection{MONSID Implementation in ROS}
The MONSID executive, including the diagnostic engine, data handling, and recording was implemented in a ROS node. Service calls are provided to start and stop the engine and ascertain current state. ROS topic messages include the health state of each modeled item, if a fault was found, and MONSID system state information (e.g., is processing enabled, number of commands received, number of command errors, and others). The health status message indicates which arm components are healthy and if any are not, which is faulty or suspected of being faulty. The system state message can be used to determine the correct functioning of the data handler and engine. The data handler, called the DataCoordinator, subscribes to four different OWLAT arm messages and creates a single time slice of sensor (message) data for the engine to consume and run through the model. The OWLAT arm messages are published at roughly 500 Hz. In order to reduce message traffic, the data was down-sampled to 50 Hz. This reduced rate was sufficient to satisfy the MONSID model fidelity. The DataCoordinator also optionally records data logs of MONSID engine operations at each processing time into files called playback. The playback files contain detailed records of MONSID health analyses. When visualized in the MONSID Toolkit app, playbacks allow users to “look under the hood” to see how a particular diagnosis was determined. Playbacks are a useful post-run analysis tool. For “live” operations during testing and demonstrations, a Toolkit feature called livestream provides a page for users to see top-level health status results as they are generated. Livestream was included in the ROS implementation. 

Assuming that the MONSID ROS node is running and the engine has been started, the nominal operation proceeds as follows. The DataCoordinator collects the robot arm messages as they are published and creates a down-sampled record of data aligned to a single time slice needed for the model. The MONSID engine waits until it receives a time slice record at which time it processes it through the MONSID model. A health state determination is made at each time slice record to provide real-time knowledge of health state. MONSID assumes that all states are healthy until proven otherwise. If a fault is found, the engine determines the faulty component(s) and updates the affected health states. At the end of the processing cycle, the DataCoordinator publishes the health status and engine state messages and logs diagnostic data (if enabled) to a playback file. The Toolkit Livestream page is updated with the latest robot arm health states. The engine then waits until a new time slice record is provided. This processing cycle continues until the engine is commanded to stop. As previously mentioned, fault injection on the testbed was limited to simulated encoder biases. During testing, biases were injected into several different joint encoders. One such run is shown in the snapshot of the Livestream page in Fig.~\ref{fig:monsid3}. A fault was injected in the sixth joint encoder (a wrist joint). MONSID detected the fault shortly after it was injected and correctly identified J6\_angle\_pos as faulty.
  
\section{RECALIBRATION} 
\label{sec:recal}
The successful collection of samples from icy moons and the efficiency of the mission depend heavily on a reliable, data-efficient, autonomous calibration process. Upon landing and initial deployment, it is crucial to recalibrate the robot arm in situ to account for any distortions that may have occurred during the multi-year space flight to the icy moons of Europa and Enceladus. These missions require a high level of automation that can withstand any disruptions to the robot and tool geometries resulting from forceful interactions with the terrain, joint freeze-ups, or improper loading of the sampling tools. Therefore, an autonomous kinematic calibration method is essential to provide the necessary accuracy for the robot arm.

To address the challenges associated with deploying robotic manipulators on future sample missions to icy moons, we present a novel, non‐parametric, and data‐efficient online calibration method that simultaneously learns the residual forward kinematics errors and optimizes the corresponding corrected Denavit-Hartenberg (DH) parameters \cite{das2024bayesian, dacs2023active, touma2023towards}. 

Our approach \cite{das2024bayesian} is built upon a geometry‐aware Bayesian optimal experimental design framework, where Gaussian Processes (GP) equipped with Riemannian Matérn kernels on \(\mathbb{S}^3\times\mathbb{R}^3\) accurately capture both the rotational and translational discrepancies along with their uncertainties. A geodesic distance–based objective function quantifies the discrepancy between the computed end-effector pose—derived from the nominal kinematics—and the noisy measurements obtained from fiducial markers. An online Gaussian Process-\textit{upper confidence bound} (GP-UCB) algorithm \cite{das2024bayesian} then dynamically selects the most informative measurement configurations, adapting to the specific error characteristics of the robot. Finally, the learned error estimates are integrated into an efficient quadratic program that updates the DH parameters, thereby jointly refining the kinematic model and achieving enhanced calibration accuracy.

\subsection{\texorpdfstring{Bayesian Optimization on ${\mathbb{S}^3 \!\times\! \mathbb{R}^3}$}{Bayesian Optimization on S3 x R3}}
\label{sec:main}
This section presents our geometry-aware Bayesian optimization framework for the experiment design problem of online kinematic calibration. 

We use standard notation: ${\mathbb{R}}$ represents the set of real numbers, and $\mathbb{N}$ represents the set of natural numbers. The Euclidean norm of a matrix is denoted by $\|\!\cdot\!\|$. ${\langle \cdot, \cdot \rangle}$ is the dot product. ${\mathbf{f} \!\sim\! \mathbf{GP} (\mu, k)}$ denotes that function $\mathbf{f}(\cdot)$ is sampled from a GP. We use the symbol $\mathbf{f}$ to represent various functions distinguished by subscripts and context.

\textbf{Forward Kinematics:} This paper focuses on serial, or open, kinematic chain that consists of ${n_j \!\in\! \mathbb{N}}$ joints. Each joint has one degree of freedom, a revolute joint with the \textit{joint variable} ${\theta_i \!\in\! [0, 2 \pi ) }$, or a prismatic joint with the joint variable ${\theta_i \!\in\! \mathbb{R} }$, connecting ${n_l \!=\! n_j \!+\! 1}$ links that form a single serial chain. Given a set of joint angles ${\uptheta \! \triangleq \! [ \theta_1, \ldots, \theta_{n_j} ]^\top \!\in\! \Theta \!\subset\! \mathbb{R}^{n_j}}$, the {\em forward kinematic} function describes the location of the \textit{end-effector} reference frame $\mathcal{T}$ relative to the \textit{base frame} $\mathcal{B}$;  ${ \mathbf{f}_{\uptheta \to \mathbf{T}} \!:\! \Theta \!\to\! \mathrm{SE(3)}}$. 

One of the most popular techniques for parameterizing the forward kinematics function is the \textit{Denavit-Hartenberg} (DH) convention.  In this convention (see \cite{murray2017} for details), {\em link frames} are defined for each link in the chain.  Let the transformation between adjacent link frames be denoted as ${ \mathbf{f}_{\theta_i \to \mathbf{T}_{i\!-\!1,i}} \!:\! \Theta_i \!\to\! \mathrm{SE(3)}}$, where ${\Theta_i \!\in\! \mathbb{S}^{1}}$ or ${\Theta_i \!\in\! \mathbb{R}}$. As a function of the link frame transformations, the forward kinematics map is given by 
\begin{equation}
\label{eq:forward}
\mathbf{f}_{\uptheta \to \mathbf{T}} \!=\!  \begin{bmatrix}
\mathbf{R}(\uptheta) & \mathbf{p}(\uptheta) \\
\mathbf{0}^\top & 1 \end{bmatrix} \!=\! \mathbf{f}_{\mathcal{B}} \Bigg ( \prod_{i=1}^{i = n_j} \mathbf{f}_{\theta_i \to \mathbf{T}_{i\!-\!1,i}} \Bigg )  \mathbf{f}_{\mathcal{T}} ,
\end{equation}
where ${\mathbf{f}_{\mathcal{B}}, \mathbf{f}_{\mathcal{T}} \!\in\! \mathrm{SE}(3)}$ are the transformation matrices for the base frame and end-effector with respect to an inertial frame, and the last link, respectively. The link transformation matrices ${ \mathbf{f}_{\theta_i \!\to\! \mathbf{T}_{i\!-\!1,i}}}$ have the form
\begin{eqnarray*}
&\mathbf{f}_{\theta_i \!\to\! \mathbf{T}_{i\!-\!1,i}}(\theta_i) = \\
&\begin{bmatrix} 
\cos{\phi_i} & \!\!-\!\sin{\phi_i} \cos{\alpha_i} & \sin{\phi_i} \sin{\alpha_i} & a_i \! \cos{\phi_i}\\ 
\sin{\phi_i} & \cos{\phi_i} \cos{\alpha_i} & \!\!-\! \cos{\phi_i} \sin{\alpha_i} & a_i \! \sin{\phi_i}\\
0 & \sin{\alpha_i} &  \cos{\alpha_i} & d_i \\
0 & 0 &  0 & 1
\end{bmatrix} \!\! ,
\end{eqnarray*}
where ${\phi_i \!\in\! \mathbb{S}^{1}}$ is the \textit{joint angle offset}, ${\alpha_i \!\in\! \mathbb{S}^{1}}$ is the \textit{twist angle}, ${a_i \!\in\! \mathbb{R} }$ is the \textit{link length}, and ${d_i \!\in\! \mathbb{R} }$ is the \textit{link offset}. And, $\phi_i$ is the joint variable if the joint is revolute, while $d_i$ is the joint variable if the joint is prismatic \cite{murray2017}. 

The similarity of two unit quaternions (which represent rotations), $\mathbf{q}_1$ and $\mathbf{q}_2$, can be measured by their {\em geodesic distance}, ${d_{\mathbb{R}^3} \!:\! \mathbb{S}^3 \!\times\! \mathbb{S}^3  \!\to\! \mathbb{R}_0^+}$,
\begin{equation}
\label{eq:quad_dist}
d_{\mathbb{S}^3}(\mathbf{q}_1, \mathbf{q}_2) = 2 \cos^{-1} \left( \left| \langle \mathbf{q}_1, \mathbf{q}_2 \rangle \right| \right),
\end{equation}
the shortest paths between the quaternions on the surface of $S^3$.  And, similar to straight lines in Euclidean space, the second derivative is zero everywhere along a geodesic. 

Note that in \eqref{eq:quad_dist}, ${\left| \langle \mathbf{q}_1, \mathbf{q}_2 \rangle \right| \!=\! \left| \langle \mathbf{q}_1, -\mathbf{q}_2 \rangle \right|}$; therefore, $\mathbf{q}_2$ and $-\mathbf{q}_2$ equivalently represent the same rotation. We thus select the shortest of the two arcs connecting $\mathbf{q}_1$ to $\mathbf{q}_2$ or $-\mathbf{q}_2$:
\begin{equation*}
 \min \!\left( 2 \cos^{-1} \!\left( \langle \mathbf{q}_1, \mathbf{q}_2 \rangle \right), 2 \cos^{-1} \! \left( \langle \mathbf{q}_1, -\mathbf{q}_2 \rangle \right) \right) \!\equiv\! d_{\mathbb{S}^3}(\mathbf{q}_1, \mathbf{q}_2).
\end{equation*}
Therefore, \eqref{eq:quad_dist} implicitly accounts for the ambiguous sign. 

\textbf{Gaussian Process:} A GP is a collection of random variables such that any finite subset of them has a joint Gaussian distribution. GPs provide a stochastic, data-driven, supervised machine learning approach to specify the relations between input and output data sets of unknown functions through Bayesian inference \cite{GP2006}. 

Let the noisy output of an unknown (black box) function 
\begin{equation}
\label{eq:unknown}
\mathbf{y} \!=\! \mathbf{f}(\mathbf{x}) \!+\! \epsilon
\end{equation}
with ${\mathbf{x} \!\in\! \mathbf{X} \!\subseteq\! \mathbb{R}^{n_{\mathbf{x}}}, \ \mathbf{f} \!:\! \mathbf{X} \!\to\! \mathbb{R}}$, be perturbed by an i.i.d zero-mean Gaussian noise ${\epsilon \!\sim\! \mathcal{N} (0, \sigma^2_{\epsilon})}$. We assume that function $\mathbf{f}$ is distributed as a GP: ${\mathbf{f}(\mathbf{x}) \!\sim\! \mathbf{GP} (\mu(\mathbf{x}), \mathbf{k}(\mathbf{x}, \mathbf{x}'))}$, with a \textit{mean function} ${\mu \!:\! \mathbf{X} \!\to\! \mathbb{R}}$ and a \textit{covariance/kernel function} ${\mathbf{k} \!:\! \mathbf{X} \!\times\! \mathbf{X} \!\to\! \mathbb{R}_0^+}$, for any ${\mathbf{x}, \mathbf{x}' \!\in\! \mathbf{X}}$. Then, the GP is fully specified as
\begin{eqnarray*}
&\mu(\mathbf{x})  \triangleq \mathop{{}\mathbb{E}} \left[ \mathbf{f}(x) \right],\\
&\mathbf{k}(\mathbf{x}, \mathbf{x}')  \triangleq \mathop{{}\mathbb{E}} \left[ (\mathbf{f}(\mathbf{x}) - \mu(\mathbf{x}) ) (\mathbf{f}(\mathbf{x}') - \mu(\mathbf{x}') ) \right].
\end{eqnarray*}

Given a collection of training data ${\mathcal{D}_n \!\triangleq\! \left\{ \mathbf{x}_i, \mathbf{y}_i \right\}_{i \!=\! 1 }^n}$ with inputs ${\mathcal{X} \!\triangleq\! \left\{ \mathbf{x}_i \right\}_{i \!=\! 1 }^n}$ and outputs ${\mathcal{Y} \!\triangleq\! \left\{ \mathbf{y}_i \!=\! \mathbf{f}(\mathbf{x}_i) \!+\! \epsilon \right\}_{i \!=\! 1 }^n}$, \textit{GP regression} predicts the output based on an input test point ${ \mathbf{x}^* \!\in\! \mathbf{X}}$. The \textit{posterior} of the GP conditioned on the observations, ${\mathbf{f}(\mathbf{x}^*) \big | \mathbf{x}^*, \!\mathcal{D}_n}$, is also a Gaussian distribution with the \textit{posterior mean} $\mu_*$ and the \textit{posterior variance} $\sigma_*^2$ given by
\begin{align}
\begin{split}
\label{eq:update}
 &\!\!\! \mu_*(\mathbf{x}^*)  \!=\! \mu(\mathbf{x}^*) \!+\! \mathbf{K}(\mathbf{x}^*, \!\mathcal{X}) \big(\Tilde{\mathbf{K}}(\mathcal{X}, \mathcal{X}) \big)^{-1} (\mathcal{Y} \!-\! \mu(\mathcal{X})), \\
  &\!\!\! \sigma_*^2(\mathbf{x}^*)  \!=\! \mathbf{k}(\mathbf{x}^*, \mathbf{x}^*) \!-\! \mathbf{K}(\mathbf{x}^*, \!\mathcal{X}) \big(\Tilde{\mathbf{K}}(\mathcal{X}, \mathcal{X}) \big)^{-1} \mathbf{K}(\mathcal{X}, \mathbf{x}^*),
\end{split}
\end{align}
where ${\Tilde{\mathbf{K}}(\mathcal{X}, \mathcal{X}) \!\triangleq\! \mathbf{K}(\mathcal{X}, \mathcal{X}) \!+\! \sigma^2_{\epsilon} \mathbf{I}}$, and ${\mathbf{K}(\mathbf{x}^*, \mathcal{X}) \!\in\! \mathbb{R}^{1 \times n}}$, ${\mathbf{K}(\mathcal{X}, \mathcal{X}) \!\in\! \mathbb{R}^{n \times n}}$, ${\mathbf{K}(\mathcal{X}, \mathbf{x}^*) \!\in\! \mathbb{R}^{n \times 1}}$ are the \textit{covariance matrices} for the set of points, which measure the correlations between the inputs with the components for all ${i, j \!\leq\! n}$: ${\left [ \mathbf{K}(\mathbf{x}^*, \mathcal{X}) \right ]_{1,j} \!=\! \mathbf{k}(\mathbf{x}^*, \mathbf{x}_j\!)}$, ${\left [ \mathbf{K}(\mathcal{X}, \mathcal{X}) \right ]_{i,j} \!=\! \mathbf{k}(\mathbf{x}_i, \!\mathbf{x}_j)}$, ${\left [ \mathbf{K}(\mathcal{X}, \mathbf{x}^*) \right ]_{j,1} \!=\! \mathbf{k}(\mathbf{x}_j, \mathbf{x}^*)}$.

One frequently used kernel function on a Euclidean space $\mathbb{R}^{n_{\mathbf{x}}}$ is the \textit{squared-exponential} (SE) kernel given by
\begin{equation*}
\mathbf{k}_{SE}(\mathbf{x}_i, \mathbf{x}_j) = \sigma^2_{f} \exp \bigg (  \dfrac{-\|\mathbf{x}_i - \mathbf{x}_j\|^2}{2 \beta^2}  \bigg ) + \sigma^2_{n} \mu(\mathbf{x}_i, \mathbf{x}_j) ,
\end{equation*}
where ${\lambda_h \!\triangleq\! [\beta~\sigma_{f}~\sigma_{n} ]^\top}$ are the tunable hyperparameters and $\mu$ is the Kronecker delta function. We remark that the SE kernel is a valid kernel for the translational motion ${\mathbf{p}}$ in ${\mathbb{R}^3}$.

\textbf{Squared-Exponential Kernel on ${\mathbb{S}^3 \!\times\! \mathbb{R}^3}$:} Valid SE and Mat\'ern kernel examples on Riemannian manifolds, based on the solutions to stochastic partial differential equations, were presented in \cite{borovitskiy2020matern}. On the sphere $\mathbb{S}^{3}$ a SE kernel is given in \cite{borovitskiy2020matern} (Example 9, Eq. 72) as
\begin{eqnarray*}
&\mathbf{k}_{\mathbb{S}^3}(\mathbf{q}_1, \mathbf{q}_2) \!=\! \\ &\!\!\!\!\!\frac{\sigma^2}{C_\infty} \sum_{n=0}^{\infty} c_{n,3} ~\! \mathcal{C}_n^{(1)} ( \cos(d_{\mathbb{S}^3}(\mathbf{q}_1, \mathbf{q}_2)) ) \exp{\left(\!\!-\frac{\kappa^2}{2} n(n \!+\! 2)\!\right)} ,
\end{eqnarray*}
where ${c_{n,3}}$ are constants, $\mathcal{C}_n^{(1)}$ are the Gegenbauer polynomials, ${\kappa \!>\! 0}$ is the length scale, ${\sigma \!>\! 0}$ is a hyperparameter to regulate the variability of the GP, and $C_\infty$ is a normalizing constant that guarantees ${\mathbf{k}_{\mathbb{S}^3}(\mathbf{q}, \mathbf{q}) \!=\! \sigma^2}$, see \cite{borovitskiy2020matern}. 

Note that while ${\mathbb{S}^3 }$ is a compact Riemannian manifold,  ${\mathbb{R}^3}$ is non-compact; thus, ${\mathbb{S}^3 \!\times\! \mathbb{R}^3}$ is non-compact. We define a product kernel on this non-compact space as the multiplication of the squared-exponential kernel for the position components and $\mathbf{k}_{\mathbb{S}^3}$ for the rotational components:
\begin{equation*}
\mathbf{k}_{\mathbb{S}^3 \!\times\! \mathbb{R}^3}(\mathbf{x}_i, \mathbf{x}_j) \triangleq \sigma^2_{s} \mathbf{k}_{\mathbb{S}^3}(\mathbf{q}_i, \mathbf{q}_j) \!~ \mathbf{k}_{SE}(\mathbf{p}_i, \mathbf{p}_j)  ,
\end{equation*}
which is a valid kernel on ${\mathbb{S}^3 \!\times\! \mathbb{R}^3}$, with ${\sigma \!>\! 0}$, since both $\mathbf{k}_{SE}$ and $\mathbf{k}_{\mathbb{S}^3}$ are positive definite. 

\textbf{Geometry-aware Bayesian Optimization:}
Bayesian optimization algorithms search for optimal input values over an unknown objective function through iterative evaluations. One first creates a probabilistic model of the function based on previous evaluations and then uses this model to select candidate optimal points for evaluation. In this study, we consider a black box function with the form: 
\begin{equation}
\label{eq:calib_cost}
\!\!\!\mathbf{f}(\mathbf{p}, \mathbf{q}) \!=\! - \left( \!\! \alpha_1 \frac{\mathbf{f}_\mathbf{p}(\mathbf{p})}{\sup_{\mathbf{p} \in \mathbb{R}^3}{|\mathbf{f}_\mathbf{p}}(\mathbf{p})|} \!+\! \alpha_2 \frac{\mathbf{f}_\mathbf{q}(\mathbf{q})}{\sup_{\mathbf{q} \in \mathbb{S}^3}{|\mathbf{f}_\mathbf{q}}(\mathbf{q})|} \! \right),
\end{equation}
where ${\mathbf{f}  \!:\! \mathbb{S}^3 \!\times\! \mathbb{R}^3  \!\to\! \mathbb{R}}$, ${\mathbf{f}_\mathbf{p} \!:\! \mathbb{R}^3  \!\to\! \mathbb{R}}$, ${\mathbf{f}_\mathbf{q} \!:\! \mathbb{S}^3  \!\to\! \mathbb{R}}$, and the terms ${ \sup_{\mathbf{p} \in \mathbb{R}^3}{|\mathbf{f}_\mathbf{p}}(\mathbf{p}) |}$ and ${\sup_{\mathbf{q} \in \mathbb{S}^3 }{|\mathbf{f}_\mathbf{q}}(\mathbf{q}) |}$ normalize a function of position and an orientation to the range ${[-1, 1]}$. The weights ${\alpha_1 \!>\! 0}$ and ${\alpha_2 \!>\! 0}$, with ${\alpha_1 \!+\! \alpha_2 \!=\! 1}$, balance the relative importance of the position and orientation errors. 

In our calibration process, the functional form of $\mathbf{f}$ in (\ref{eq:calib_cost}) is unknown, but it is assumed to satisfy \eqref{eq:unknown} and \eqref{eq:calib_cost}. Pointwise values of these functions can be sampled via noisy measurements. In our approach, Bayesian optimization utilizes these measurements to solve the following optimization problem: 
\begin{equation*}
    {\mathbf{p}^*, \mathbf{q}^*= \ } 
\argmax_{(\mathbf{\mathbf{q}, \mathbf{p}}) \in (\mathbb{S}^3 \!\times\! \mathbb{R}^3)}~  \mathbf{f}(\mathbf{\mathbf{p}, \mathbf{q}}) ,
\end{equation*}
by representing $\mathbf{f}$ with a GP: ${\mathbf{f}(\mathbf{x}) \!\sim\! \mathbf{GP}  (\mu(\mathbf{x}), \mathbf{k}(\mathbf{x}, \mathbf{x}'))}$ for ${\mathbf{x} \!=\! (\mathbf{\mathbf{p}, \mathbf{q}})}$, ${\mathbf{x}' \!=\! (\mathbf{\mathbf{p}', \mathbf{q}'})}$, ${\mathbf{x}, \mathbf{x}' \!\in\! \mathbb{S}^3 \!\times\! \mathbb{R}^3}$. This approach offers the significant advantage that predictive uncertainty quantification can guide the trade-off between exploration and exploitation. This balance is achieved by using a well-designed decision rule to choose optimal actions $\mathbf{x}^*$.   Careful algorithm design can produce sample-efficient algorithms, and therefore rapid in-the-field recalibration of a manipulator.    

In Bayesian optimization, one refines a belief about $\mathbf{f}$ at each observation through a Bayesian posterior update. A \textit{utility function} (\textit{acquisition function}) ${V_k \!:\!  \left ( \mathbb{S}^3 \!\times\! \mathbb{R}^3 \right ) \!\to\! \mathbb{R}}$ guides the search for the optimal point. This function evaluates the usefulness of candidate points for the next function evaluation. The utility function is maximized, based on the available data, to select the next query point.  Finally, after a certain number of queries, the algorithm recommends that the point is the best estimate of the optimum. In this paper, we use the following GP-UCB decision algorithm \cite{srinivas2009} to determine the next sampling point based on currently available measurements:
\begin{equation*}
{\mathbf{x}_{k}= \ } 
\argmax_{\mathbf{x} \in (\mathbb{S}^3 \!\times\! \mathbb{R}^3)}~  \big (  \mu_{k-1}(\mathbf{x}) + \sqrt{\beta_k}\!~ \sigma_{k-1}(\mathbf{x}) \triangleq V_k(\mathbf{x}) \big ) ,
\end{equation*}
where ${\beta_k \!>\! 0 }$ is an iteration-varying parameter that weights the uncertainty in the selection of the next sampling point.

\subsection{Experimental Design for Calibration}
 
This section presents our geometry-aware Bayesian optimization framework for the experiment design problem of online kinematic calibration. Our goal is to search for a sequence of experimental measurement poses  $\mathcal{D}_n \!\triangleq\! \left\{ \mathbf{x}_i, \mathbf{y}_i \right\}_{i \!=\! 1 }^n$, ${\left\{\mathbf{x}_i \!\in\! (\mathbb{S}^3 \!\times\! \mathbb{R}^3)\right\}_{i \!=\! 1 }^n \!\triangleq\! \mathcal{I}_n}$, ${\mathbf{y}_i \!\in\! \mathbb{R}}$ that quickly recalibrate the arm. We remark that since the joint variables are assumed to be uncertain after a miscalibration is detected, we need another input argument for the black box objective function. Therefore, we seek optimal end-effector configurations instead of optimal joint variables. 

Assume that the forward kinematics errors can be completely captured by time-invariant DH parameter errors ${\Bar{\delta} \!\triangleq\! [\delta \Bar{\phi}\ \delta \Bar{\alpha}\  \delta \Bar{a}\  \delta \Bar{d}]^\top \!\in\! \mathbb{R}^{4n_j}}$.  For known DH parameters and given joint variables, the value of the \textit{computed end-effector pose} is denoted as ${ [\tilde{\mathbf{q}}~\tilde{\mathbf{p}}]^\top \!=\!\mathbf{f}_{\uptheta \to \mathbf{T}}(\Psi) }$, where ${\Psi \!\triangleq\! [\Bar{\phi}~\Bar{\alpha}~\Bar{a}~\Bar{d}]^\top \!\in\! \mathbb{R}^{4n_j}}$ are the \textit{nominal} DH parameters.  However, under parameter errors $\Bar{\delta}$, the \textit{measured (actual) end-effector pose} value, ${ [{\mathbf{q}}~{\mathbf{p}}]^\top \!=\!\mathbf{f}_{\uptheta \to \mathbf{T}}(\Psi \!+\!\Bar{\delta} ) }$, might be different than the computed ones.
We represent the forward kinematic error as:
\begin{equation*}
  [{\mathbf{q}}~{\mathbf{p}}]^\top - [\tilde{\mathbf{q}}~\tilde{\mathbf{p}}]^\top = \mathbf{f}_{\uptheta \to \mathbf{T}}(\Psi +\Bar{\delta} ) \!-\! \mathbf{f}_{\uptheta \to \mathbf{T}}(\Psi) \triangleq \Delta({\mathbf{q}}~{\mathbf{p}}),
\end{equation*}
where ${\Delta}$ maps ${\left ( \mathbb{S}^3 \!\times\! \mathbb{R}^3 \right )}$ to ${  \left ( \mathbb{S}^3 \!\times\! \mathbb{R}^3 \right )}$. However, in order to utilize Bayesian optimization, we need a function that maps ${\left ( \mathbb{S}^3 \!\times\! \mathbb{R}^3 \right )}$ to ${\mathbb{R}}$ as given in \eqref{eq:calib_cost}. In this study, we use 
\begin{equation}
\label{eq:objective}
\mathbf{f}_\mathbf{p}(\mathbf{p}) \!=\!   \|  \mathbf{p} - \tilde{\mathbf{p}} \|, \quad  \mathbf{f}_\mathbf{q}(\mathbf{q}) \!=\! d_{\mathbb{S}^3}(\mathbf{q}, \tilde{\mathbf{q}}),
\end{equation} in \eqref{eq:calib_cost} to represent the unknown cost function.

In the implementation of our approach, we use vision-guided manipulation techniques similar to those used in planetary robotics \cite{bajracharya2007} position control frameworks to navigate the robotic arm to specific test locations using a depth camera system. This visual servoing based approach enables accurate end-effector placement despite miscalibration. We employ fiducial markers on the end-effector for high precision, and a camera detects the marker to correct the arm poses (see Fig.~\ref{fig:owlat_tom_edit}).

Pseudo-code of the proposed Bayesian optimal experimental design for the kinematic calibration is presented in Algorithm~\ref{alg1}. This algorithm implements an online geometry-aware Bayesian optimization to select the sequence of end-effector test locations that will quickly and best recalibrate the manipulator. It leverages the geometry of the special Euclidean group to optimize robot pose selection. The algorithm iteratively selects optimal end-effector poses by maximizing an acquisition function based on the GP model of the kinematic error. For each iteration, the robot is positioned to the chosen pose, and joint variables are measured. The algorithm then computes the expected pose using nominal DH parameters and compares it to the actual measured pose. This comparison yields position and orientation errors, which are used to update the objective function. The GP model is subsequently updated using these new noisy observations, refining the understanding of the kinematic error space. This process continues for a specified number of iterations, resulting in a set of optimally selected poses for kinematic calibration. The algorithm's strength lies in its ability to balance exploration and exploitation of the error space, guided by the GP model's mean and variance predictions.
\begin{algorithm}[tb]
 \caption{Geometry-aware Bayesian optimization for  kinematic calibration experimental design}
 \label{alg1}
 \begin{algorithmic}
 \renewcommand{\algorithmicrequire}{\textbf{Input:}}
 \renewcommand{\algorithmicensure}{\textbf{Output:}}
 \REQUIRE Input set $\mathbf{X}$: ${\mathbf{X} \!\subset\!  (\mathbb{S}^3 \!\times\! \mathbb{R}^3)}$ \\~~~
 Nominal DH parameters ${\Psi}$ \\~~~
 Forward kinematics ${\mathbf{f}_{\uptheta \to \mathbf{T}}(\Psi)}$ \\~~~
 Kernel function $\mathbf{k}_{\mathbb{S}^3 \!\times\! \mathbb{R}^3}$ \\~~~
 GP prior $\mu(\mathbf{x}),~ \sigma^2(\mathbf{x})$ \\~~~
 Parameter $\beta_k$ \\~~~
 Weights ${\alpha_1}$, ${\alpha_2}$, ${\gamma_1}$, ${\gamma_2 \!>\! 0}$
 \ENSURE  ${\mathcal{I}_n \!\triangleq\! \left\{ \mathbf{x}_i \right\}_{i \!=\! 1 }^n}$, ${\mathbf{x}_i \!\in\! (\mathbb{S}^3 \!\times\! \mathbb{R}^3)}$, ${\mathcal{F}_n \!\triangleq\! \left\{ \tilde{\mathbf{x}}_i \right\}_{i \!=\! 1 }^n}$
  \FOR {$i = 1, 2, \ldots, n$,}
  \STATE Choose $ {\mathbf{x}_i^*= \ }
\argmax_{\mathbf{x} \in \mathbb{S}^3 \!\times\! \mathbb{R}^3} ~ V_k(\mathbf{x})$
  \STATE Orient the end-effector to the goal pose $\mathbf{x}_i^*$
  \STATE Measure the joint variables $\uptheta_i$
  \STATE Compute the pose ${\tilde{\mathbf{x}}_i = [\tilde{\mathbf{q}}_i~\tilde{\mathbf{p}}_i]^\top }$ with $\uptheta_i$: ${\mathcal{F}_n}$
  \STATE Obtain the unit quaternion with the mapping $\mathbf{f}_{  \mathbf{R} \to \mathbf{q}}$
  \STATE Compute the values of $\mathbf{f}_\mathbf{p}$ and $\mathbf{f}_\mathbf{q}$ in \eqref{eq:objective}
  \STATE Compute the value of the objective function $\mathbf{f}$ in \eqref{eq:calib_cost}
  \STATE Update $\mu_{i}$ and $\sigma_{i}$ via Equation \eqref{eq:update}
  \ENDFOR
 \end{algorithmic} 
 \end{algorithm}

\subsection{Optimization-Based Kinematic Calibration}
\label{sec:para}
Data from Algorithm~\ref{alg1} drives the kinematic calibration process, which minimizes the discrepancy between the observed end-effector data and the calculated forward kinematics data:
\begin{equation}
\label{eq:OptMea}
 \Delta_n = [\mathcal{I}_n]^\top - [\mathcal{F}_n]^\top \in \mathbb{R}^{7n},
\end{equation}
where $n$ is the number of measurements, $\mathcal{I}_n$ represents the measured (actual) pose values, $\mathcal{F}_n$ is the computed end-effector pose, and these data sets are defined in Algorithm~\ref{alg1}. 
\begin{remark}
\label{re:quat_data}
We remark that to compute the difference between the unit quaternion components of $\mathcal{I}_n$ and $\mathcal{F}_n$, we use distance formula  \eqref{eq:quad_dist}. We can find the sign of the closest quaternion via the minimum operation: ${ \min \!\left( 2 \cos^{-1} \!\left( \langle \mathbf{q}_1, \mathbf{q}_2 \rangle \right), 2 \cos^{-1} \! \left( \langle \mathbf{q}_1, -\mathbf{q}_2 \rangle \right) \right)}$. Then, we obtain the quaternion difference with respect to the closest one. 
\end{remark}

Suppose that a lower bound ${\Bar{\delta}_{lb} \in \mathbb{R}^{4n_j}}$ and an upper bound ${\Bar{\delta}_{ub} \in \mathbb{R}^{4n_j}}$ for the unknown DH parameters are given, and ${\rank(\mathbf{J}_n) \!=\! 4n_j}$. Then, the linearization-based calibration problem can be solved via the following QP:
\begin{align}
{\label{eq:Cal-QP}}
\begin{array}{l}
{\Bar{\delta}^*= \ }
\displaystyle  \argmin_{\Bar{\delta} \in \mathbb{R}^{4n_j}} \ \ \ {\|\Delta_n - \mathbf{J}_n \Bar{\delta}   \|^2}  \\ [1mm]
~~~~~~~~~~~~\textrm{s.t.} ~~~~~~~~~ \Bar{\delta}_{lb} \leq \Bar{\delta}  \leq \Bar{\delta}_{ub} .
\end{array}
\end{align}
Then, we obtain ${\Psi^* \!=\! \Psi \!+\! \Bar{\delta}^*}$, and this QP is iteratively solved until an acceptable error convergence is achieved.

\begin{remark}
\label{re:jacob}
If some DH parameters are known, the Jacobian matrix $\mathbf{J}$ should be defined by the partial derivatives with respect to the unknown parameters only. Similarly, the QP is constructed and only solved for the unknown parameters. 
\end{remark}

\section{TESTBED AND HARDWARE}
\label{sec:owlat}
\subsection{OWLAT Testbed}
For experimentation with a hardware system the OWLAT testbed \cite{tevere2024owlat} was used. This testbed was designed to mimic a lander performing sampling operations on another celestial body to aid in developing robust autonomy approaches for sampling and system level operations. Experimentation with this physical system further provides support for the algorithm's ability to adjust to the noise introduced and faults that occur when working with a real-time system. In the following section, we will introduce the testbed and its capabilities and expand upon specific capabilities that were leveraged to complete the experimentation.
\begin{figure}[t]
    \centering
\includegraphics[width=1\linewidth]{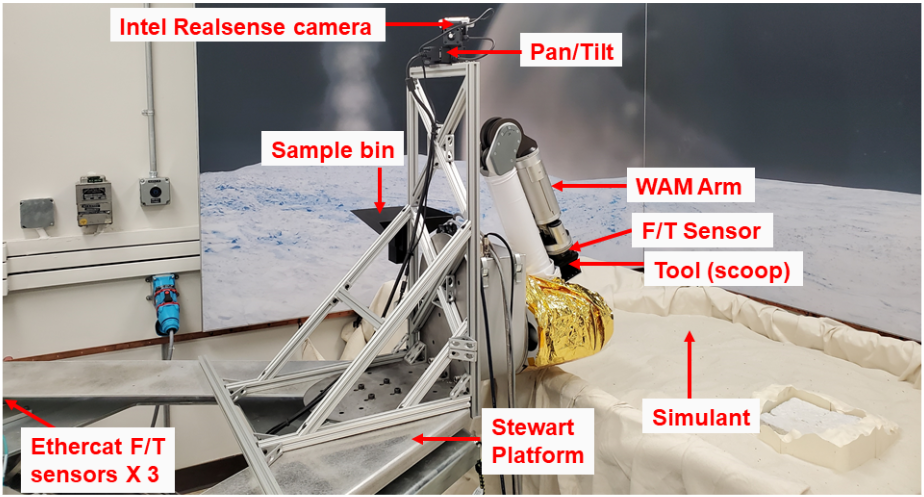}
    \caption{ The OWLAT testbed modules including a robotic arm, a Stewart platform, a mounted vision system, geotechnical and sampling tools, and a simulant workspace.
    }
   \label{fig:owlat1}
\end{figure} 

\begin{figure}[t]
    \centering
\includegraphics[width=1\linewidth]{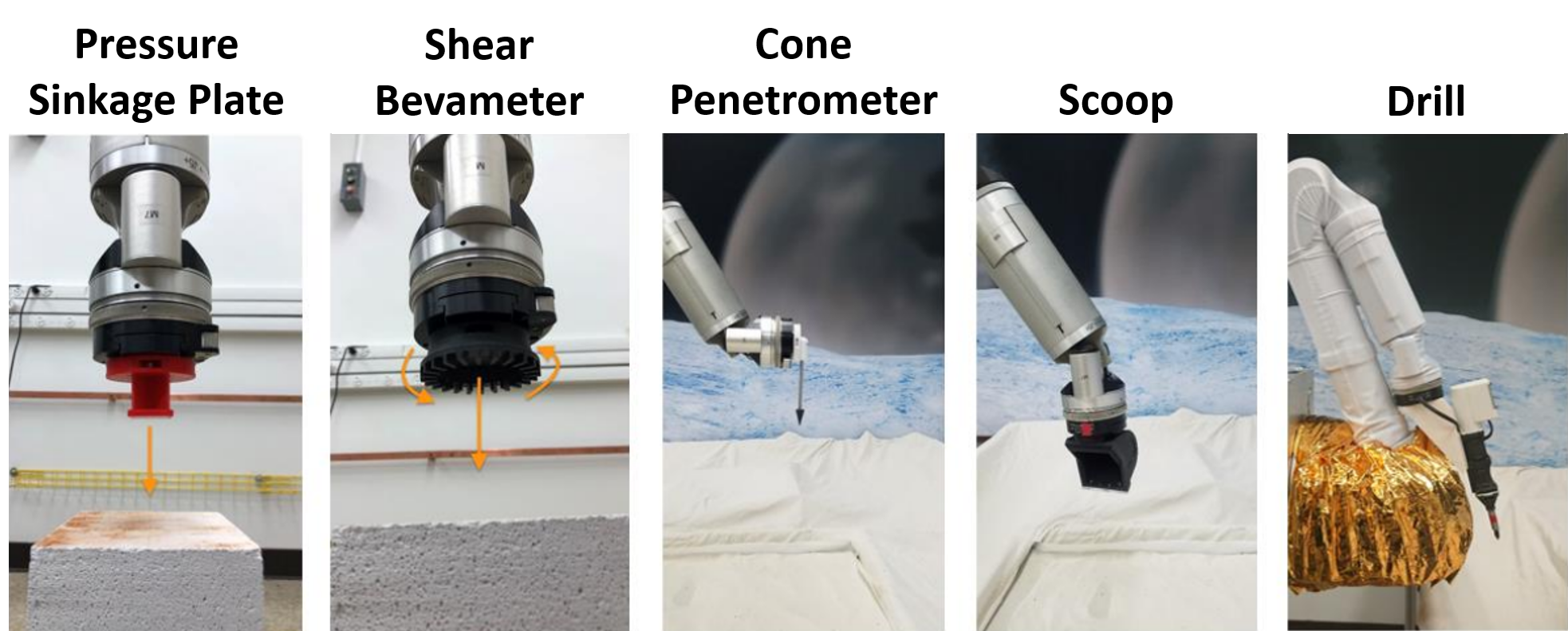}
    \caption{ The various tools available for use on the OWLAT testbed. Geotechnical tools include a pressure sinkage plate, a shear bevameter, and a cone penetrometer. Sampling tools include a scoop and a swarf-collecting drill. In this experimentation, the cone penetrometer and scoop were used.
    }
   \label{fig:owlat2}
\end{figure}

The OWLAT testbed, pictured in Fig.~\ref{fig:owlat1}, is made up of several modules working together to provide the full functionalities of the system. The system includes a 7 degrees of freedom (DoF) robotic arm, a 6 DoF lander emulation system, a 2 DoF vision system, a suite of geotechnical and sampling tools, and a simulant workspace with simulants of varying material properties. The robotic arm is a 7 DoF WAM arm made by Barrett Technology, which can be controlled with either position or torque references. A six-axis force-torque sensor at the tool interface provides feedback on interactions of the tool in the environment. The geotechnical tools that can be used include a cone penetrometer, a shear bevameter, and a pressure sinkage plate. These tools allow users to probe potential sampling targets to gain information on science relevant material properties such as bearing capacity of the ground or shear strength of a material. The sampling tools provided are a variety of scoops and a swarf-collecting drill. In this work the tools used were the cone penetrometer and scoop seen in Fig.~\ref{fig:owlat2}.

In parallel with the arm a 6 DoF Stewart platform is used to account for the interactions of the lander base with the landing surface. The platform can be controlled to simulate the behavior of a lander in low-gravity environments like Europa and Enceladus. Force-torque sensors are installed between the Stewart platform and the mounting plate for lander peripherals to capture the dynamic loads generated by the relative motion of lander components and the interaction of the sampling system with the surface. The platform can then feed these measured loads into a lander dynamics model, which accounts for parameters of the lander such as lander mass and number of legs. The expected response of the lander is then effectuated on the hardware. 

\begin{figure}[t]
    \centering
\includegraphics[width=1\linewidth]{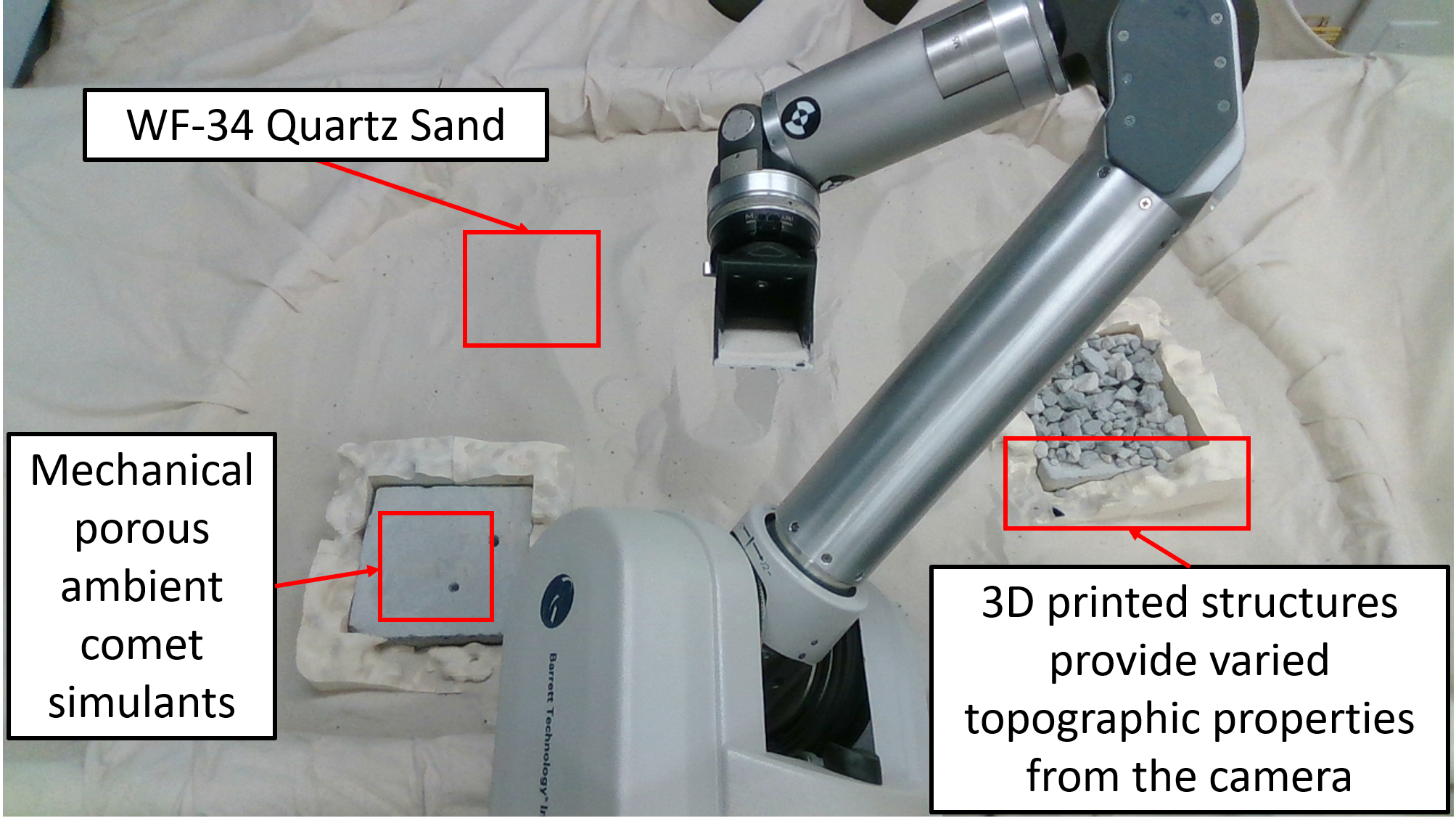}
    \caption{ An image taken from the camera on the mast shows the diversity in material and visual properties of the simulant workspace.
    }
   \label{fig:owlat3}
\end{figure}

For visual inputs the system has a singular camera on a mast placed consistent with preliminary designs for mission concepts of interest such as Europa Lander \cite{hand2022science}. While the camera is placed in an ideal location for observing interactions of the system with the simulant workspace, expected vision challenges like occlusion are still present as observed in Fig.~\ref{fig:owlat3}. The simulant workspace provides 1m$^2$ of sampling area containing diversity in material and visual properties. One simulant in the workspace is WF-34 Quartz sand. The nearly cohesionless sand was chosen to represent condensed ice that can be found coming from plumes such as those on Enceladus. The second type of simulant, mechanical porous ambient comet simulants (MPACS), represents ambient solid sintered material that can be found on the surface of Europa. The simulant is supplemented with features in the workspace to introduce some of the topographic diversity likely to exist on the surface of these planetary bodies of interest.

\begin{figure}[t]
    \centering
\includegraphics[width=1\linewidth]{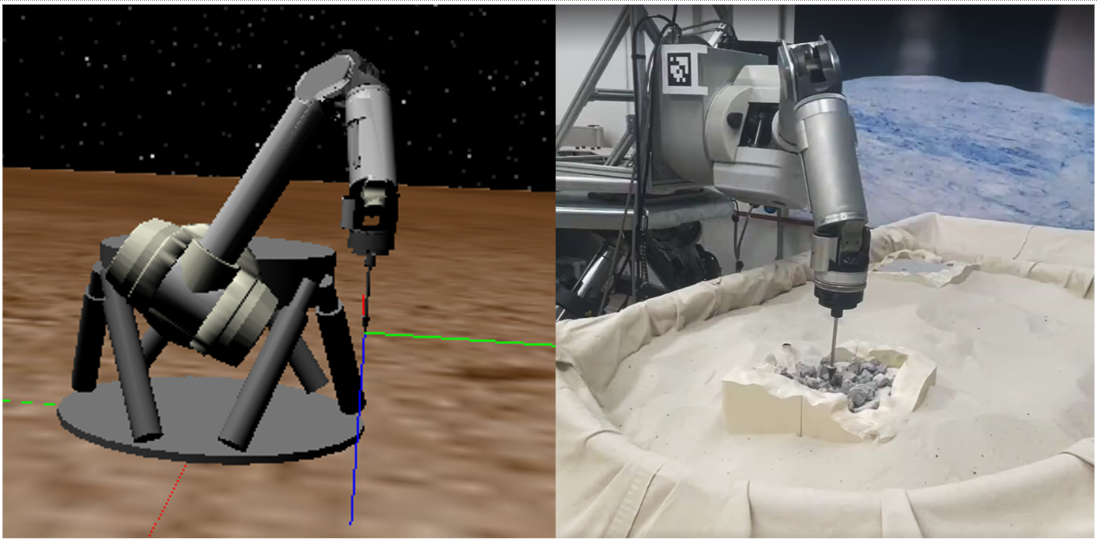}
    \caption{ A cone penetrometer test is run in the simulated testbed (left). The same test is later performed on the hardware testbed (right). The existence of both allows users to more quickly develop and deploy new features on the more complex hardware platform.
    }
   \label{fig:owlat4}
\end{figure}

To aid in the development of algorithms a simulator version of the testbed is provided. The simulator leverages the DARTS physics engine \cite{jain2020darts} and represents a majority of the subsystems and functionalities available on the hardware testbed. The simulator provides feedback modeled after the sensors used on the system, including force-torque sensor values, system and subsystem faults, and camera images. Users interact with this simulator testbed the same way they would interact with the hardware testbed, making the transition from interfacing with the simulated testbed to hardware testbed seamless. The availability of the simulated testbed can provide a platform to do higher risk development and testing. This allows users to collect initial verification of expected testbed performance as seen in Fig.~\ref{fig:owlat4}, and more safely deploy autonomy algorithms onto the physical hardware as depicted in Section~\ref{sec:recal}.

The OWLAT testbed leverages the aforementioned hardware and implements an abstracted software system to make interactions with it representative of a flight-like spacecraft. At a high level the user interacts with the testbed by sending actions and reading telemetry, much like how a rover may be commanded to drive a path on Mars. At a low level the software implements the capability to take in these actions and execute them on the testbed. Commands can include things as simple as moving the arm to a specific set of joint poses to commanding the testbed to deliver a sample to a defined location in 3D space. The software also implements fault monitoring at both a subsystem and system level to safe-guard the testbed and to provide realistic challenges of autonomous operations on a spacecraft. 

On top of basic functionality to control the testbed, some advanced capabilities are implemented in the software to introduce additional challenges to users. These capabilities, while integrated into the low-level software of the testbed, are not directly exposed to users and as such autonomy approaches must find ways to identify and overcome unexpected behaviors that occur. One of these advanced capabilities is the ability to introduce encoder errors in the system. The introduction of encoder errors can be representative of multiple types of mechanical and electrical faults that can occur within the robotic arm such as the arm falling out of calibration over time. When injected, the feedback of the joint position remains the same but the joint is physically moved to be offset by a specified account on the hardware. This fault, which propagates through a serial manipulator chain can cause abstracted failures like the cone penetrometer never interacting with the surface when a specific 3D point on the surface was specified for the command. This feature was used heavily in experimentation with the recalibration algorithm described in Section~\ref{sec:recal}.

\subsection{AI Space Cortex Hardware} 
The AI Space Cortex was deployed on a compact yet capable hardware platform, ensuring real-time and energy-efficient processing capabilities. The system ran on an Intel NUC with an 8-core CPU, 64GB of RAM, and an NVIDIA RTX 2060 (6GB VRAM) laptop GPU. However, our tests showed that the full 64GB of RAM was unnecessary. 8GB was sufficient for stable operations. The 6GB of VRAM was critical for onboard processing of foundation vision models, particularly for executing SAM-1 segmentation without offloading to external servers.

For depth perception and RGB capture, we utilized an Intel RealSense D415 camera, which provided precise spatial awareness for terrain analysis and sample targeting. The entire hardware stack can be replicated on alternative embedded platforms such as Snapdragon-similar boards or NVIDIA Jetson modules. This flexibility ensures that the AI Space Cortex can be adapted for future space missions, whether running on low-power planetary rovers or high-performance orbital platforms.

\section{RESULTS}
\label{sec:results} 
\subsection{Overview of Experimental Testing}
To evaluate the performance of the AI Space Cortex and its integrated subsystems, we conducted a series of controlled test runs under simulated mission conditions. These tests were designed to assess the system’s ability to autonomously execute science missions, detect and recover from faults, and recalibrate itself in real-time.

This section presents two main test scenarios that were repeated multiple times with similar results: 

Test 1 involved a fault injection scenario in the robotic arm during a sample delivery mission, requiring in-situ recalibration before completion. The system’s ability to detect, diagnose, and autonomously correct the kinematic discrepancy was examined.

Test 2 evaluated the AI Space Cortex’s ability to autonomously execute a complete sample collection and delivery mission. The system was tasked with segmenting the environment, identifying viable sampling sites using the Intelligent Scene Interaction (ISI) module and large language model (LLM) reasoning, validating material suitability through force-torque probing, and performing a precision-controlled scoop and sample cache delivery. The objective was to assess the Cortex’s real-time decision-making capabilities, system responsiveness, and overall efficiency in an end-to-end, AI-driven science operation.

\subsection{Test 1: Fault-Detection, Recalibration and Fault Recovery During a Sample Delivery Mission}
\textbf{Test Objectives:} The primary objectives of this test were to evaluate the AI Space Cortex’s ability to (1) detect kinematic faults in real-time, (2) autonomously correct joint misalignments through the Recalibration Algorithm, and (3) successfully complete a sample delivery mission despite the induced fault. The experiment also assessed the system’s decision-making logic in determining whether recalibration was necessary before resuming science operations.

\textbf{Experimental Setup:}
At the beginning of the test, the robotic arm was initialized in a stowed position, mimicking post-landing conditions. The system executed a full preliminary check, ensuring power stability, confirming no recent shutdown anomalies, and verifying sensor calibration, including perception systems. All ROS nodes were successfully launched, and system health indicators confirmed nominal conditions. The AI Space Cortex was set to operate in Scientific Curiosity Mode, prioritizing exploratory decision-making.

\textbf{Test Execution and Fault Injection:}
After completing its initial diagnostics, the AI Space Cortex issued an unstow command to transition the arm into an operational stance. The motion was executed successfully, and MONSID reported no discrepancies in the system's telemetry. The robotic arm proceeded to execute its first scoop attempt.

At this stage, a 0.5235 radian fault was manually injected into joint 7 (scoop wrist) via the OWLAT Testbed panel, simulating a kinematic misalignment due to mechanical degradation. Although no visible abnormality was initially observed, MONSID detected a discrepancy between the joint encoder reading and the expected mathematical model of the arm’s kinematics. The detected fault was relayed through the fault bridge to the HC, which issued an immediate halt command to prevent further deviation.

The abrupt system halt triggered an internal controller fault, which the HC’s fault wiper logic cleared to maintain system stability. However, the induced error remained unobservable in the arm’s resting position. To manifest the fault visually, an additional unstow command was issued manually, revealing a noticeable misalignment in the end-effector position. At this point, the AI Space Cortex determined that the arm could not safely proceed with the sample mission and engaged the Recalibration Algorithm.

\begin{figure}[t]
    \centering
\includegraphics[width=1\linewidth]{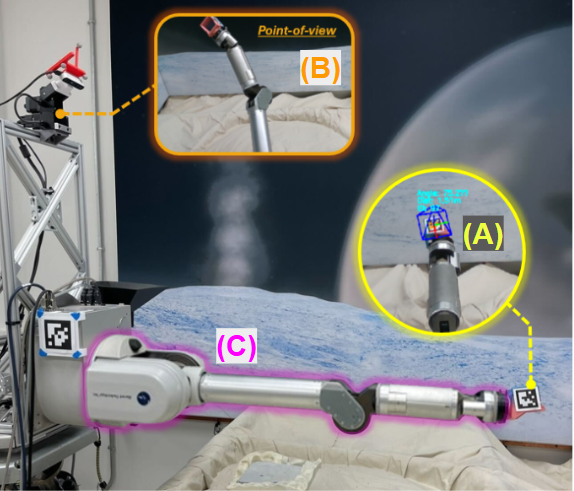}
    \caption{The OWLAT Testbed at the NASA Jet Propulsion Laboratory (JPL). Yellow inset (A)) custom scoop with 36h11 AprilTags installed for precise real-time localization of the end-effector's translation and rotations; \textit{(orange inset (B))} point-of-view from the lander’s vision system of the WAM arm in one of its other various configurations used for calibration; \textit{(purple inset (C))} segmented highlight of the WAM arm in one of its calibration poses where it is extended with a visible rotational injected encoder bias in joint-7.}
   \label{fig:owlat_tom_edit}
\end{figure}

\textbf{Recalibration Procedure:}
The AI Space Cortex activated the active learning-based recalibration process to correct the arm misalignment. Near the 7th joint, we attached a custom-designed scooping end-effector, which includes two 36h11 AprilTags \cite{apriltags}, which are used to localize the end-effector relative to the arm base, as seen in Fig.~\ref{fig:owlat_tom_edit}. The recalibration followed a structured sequence of iterative error minimization, utilizing camera-based transform estimation via 36H11 April Tags and real-time kinematic feedback.
\begin{figure}[t]
    \centering
\includegraphics[width=1\linewidth]{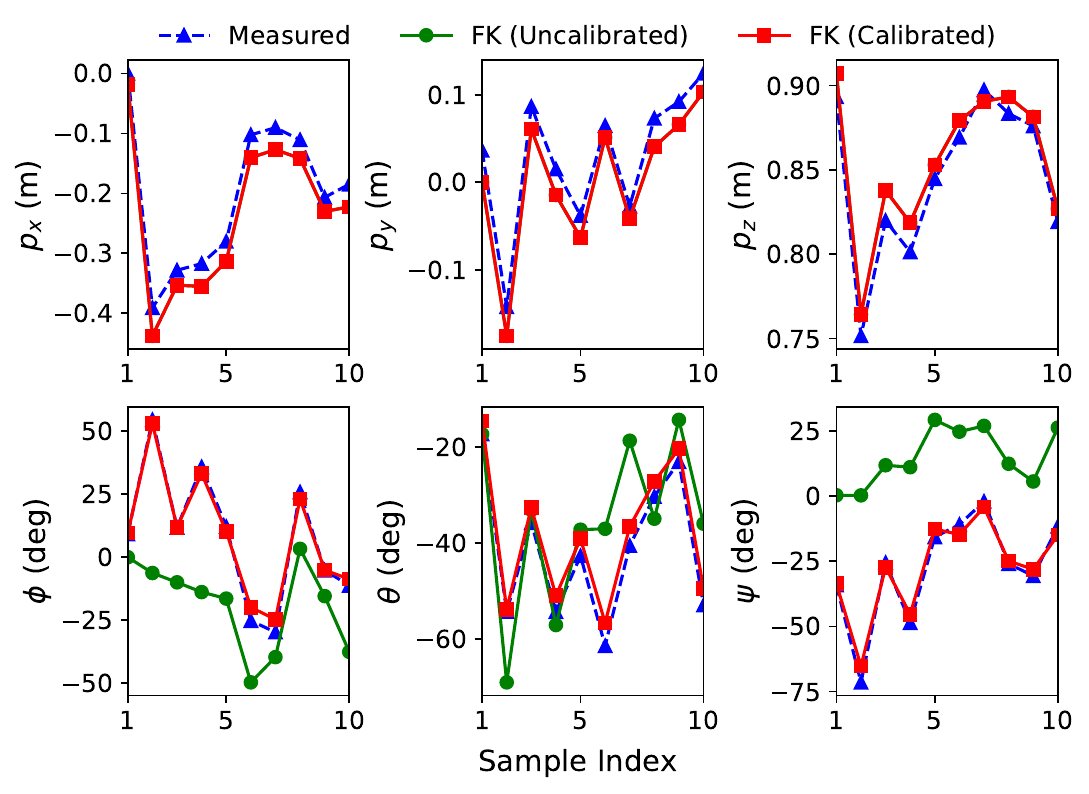}
    \caption{Kinematic calibration validation on ten distinct test measurement configurations of NASA JPL’s OWLAT.
The top row shows measured (blue triangles), uncalibrated (green circles), and calibrated (red squares) end‐effector positions $\{p_x,p_y,p_z\}$ (in meters), while the bottom row shows the Euler angles $\{\phi,\theta,\psi\}$ (in degrees).
A nominal $0.5235\,\mathrm{rad}$ bias was introduced in joint~7 before calibration. Using our \emph{geometry‐aware Bayesian optimization} procedure, the robot adaptively selected measurement poses to maximize calibration efficacy, after which a QP solver determined a final $0.57\,\mathrm{rad}$ offset. Because this bias primarily affects orientation, the uncalibrated and calibrated $\{p_x,p_y,p_z\}$ traces appear identical. Once calibrated, the forward‐kinematics errors are significantly reduced, yielding close alignment with the measured data across all ten tested poses.
Minor discrepancies arise from the marker‐based pose‐estimation process.}
   \label{fig:mainres}
\end{figure}

The system executed 38 distinct arm configurations, systematically adjusting joint positions while computing real-time transform matrices through the mast camera’s depth data. Each iteration provided updated end-effector pose estimates, which were compared against the expected kinematic model. After 38 iterations, active movement was halted, and the algorithm performed additional computational refinements without further motion.

The final computed joint offset correction was determined to be: ${\text{Accuracy} 
= 100\% \times \frac{0.52}{0.57} 
\approx 91.2\%}$. This correction achieved an accuracy improvement of 91.2\%, effectively restoring the arm to operational status. Once validated, the AI Space Cortex resumed mission execution, reissuing the scoop command to complete the sample collection process.

The result of the experiments is presented in Fig.~\ref{fig:mainres}. This figure illustrates that the proposed method can calibrate the arm with 38 pose measurements with high accuracy, thus overcoming the initially injected biases.

\textbf{Mission Completion and Performance Analysis:}
Following the recalibration process, the robotic arm successfully executed the scoop maneuver and delivered the sample to the designated cache, marking the successful completion of the mission. The recalibration phase lasted 31 minutes, and the entire mission, including fault detection, recovery, and sample delivery, was completed in 39 minutes.

These results confirm the AI Space Cortex’s ability to detect, diagnose, and correct kinematic anomalies in situ, ensuring that scientific operations can continue without direct human oversight. The successful execution of this mission highlights the system’s resilience and adaptability, demonstrating its capability to maintain autonomous function even in the presence of mechanical disruptions.

\subsection{Test 2: Successful Autonomous Sample Collection and Delivery Using the AI Space Cortex}

\textbf{Test Objectives:}
The primary objective of this test was to assess the fully autonomous operation of the AI Space Cortex, including Intelligent Scene Interaction (ISI), the Hierarchical Controller (HC), and the Explanation Engine (EE), in executing a science mission from start to completion. This included evaluating the Cortex’s ability to segment and analyze the environment, determine viable sampling sites using LLM reasoning and vision foundational models, perform material validation through force-torque probing, and execute a precision-controlled sample collection and delivery operation. Performance was measured in terms of execution time, system responsiveness, and decision-making accuracy.

\textbf{Experimental Setup:}
The test was conducted using the same hardware as in Test 1.

All ROS nodes were launched successfully, with system diagnostics confirming nominal conditions. Preliminary checks, including power validation, previous shutdown status, and sensor health, were completed in under 10 seconds. The AI Space Cortex, operating in Scientific Curiosity Mode, determined that conditions were optimal for initiating a science mission.

\textbf{Scene Segmentation and Object Selection:}
Upon initiating the science mission, the ISI module activated the mast camera, positioning it to capture the full reachable workspace of the robotic arm. The system captured a single RGB-D frame, which was immediately processed by the SAM-1 vision transformer model to generate a segmented representation of the environment. Each unique object was assigned an independent segmentation mask, and this data was transmitted to the EE for visualization.

Using ISI’s image engineering techniques, the segmentation results were analyzed to identify regions of scientific interest. The EE displayed a visualization containing all unique segmentations, as well as a filtered subset highlighting the most probable scientifically relevant targets. This process, from camera positioning to segmentation to visualization, was completed in under 3 seconds.

Following segmentation, the LLM-based reasoning module evaluated each candidate site to determine its suitability for sampling. Each segmented region was assigned a confidence integer (1–10) by the ISI module based on its estimated scoopability, material hardness, and scientific value. In this instance, seven potential sites were analyzed by the AI Space Cortex.

Three sites were deemed non-viable, with confidence values ranging from 1 to 3. Two sites were assigned medium confidence, indicating possible suitability, pending force-torque validation. One site was classified as high-priority, exhibiting low-to-medium density, absence of large obstructions, and surface resemblance to fine soil. This site was assigned a confidence integer of 8, indicating a high likelihood of successful material retrieval.

The EE visually represented these classifications in real-time, overlaying medium-confidence and high-priority sites on the RGB-D image with yellow hit markers and orange crosses, denoting areas requiring further probing validation. The EE also provided the raw Cartesian coordinate arrays for each potential sample site, aligned to the robotic arm’s base\_link reference frame, offering precise spatial mapping for subsequent operations.

Each LLM evaluation took approximately 2.5 seconds, ensuring that the site selection process was both fast and computationally efficient.

\begin{figure}[t]
    \centering
\includegraphics[width=1\linewidth]{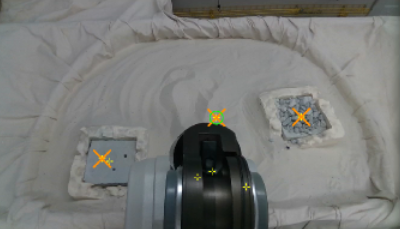}
    \caption{ The final selection marked with a green circle overlay on the RGB image. 
    }
   \label{fig:sampling_overlay}
\end{figure}

\textbf{Force-Torque Probing and Final Site Selection:} Following the ISI analysis, the Hierarchical Controller (HC) commanded the robotic arm to initiate force-torque probing at all medium-confidence and high-priority locations. The arm unstowed and executed a sequential probing routine, recording force-torque feedback at each candidate site. The EE displayed live extrapolated penetration depths (in meters), which were recorded at each site of interest: $\left[ 0.067 ~ 0 ~ 0 ~ 0.0014 ~ 0 ~ 0.0017 ~ 0
\right ] $. These values corresponded to the seven interesting sites identified by ISI. Zero values indicated virtually no penetration. Based on these measurements, \textbf{Site 1} (0.067 m) exhibited the most favorable penetration characteristics for sampling. The EE dynamically updated its visualization, marking the final selection with a green circle overlay on the RGB-D image (see Figure~\ref{fig:sampling_overlay}), signifying a confirmed viable sampling location. 

The probing sequence spanned 230 seconds. This included all force data collection, processing, and final site selection confirmation.

\textbf{Sample Collection and Delivery Execution:} With the final sampling site confirmed, the AI Space Cortex transitioned from force-torque probing to sample retrieval.

The robotic arm then executed its scoop maneuver, guided by the final site selection coordinates. The scoop operation and subsequent cache delivery were completed in approximately 124 seconds. The robotic arm then returned to a neutral position, signaling mission completion and readiness for subsequent tasks.

\textbf{Performance Analysis and System Metrics:} The AI Space Cortex successfully executed the entire sampling mission, from initial scene segmentation to final sample delivery, in approximately 8 minutes and 40 seconds (approx. 520 seconds). The full pipeline can be seen in Fig~\ref{fig:figureA2}. Throughout the mission, the system operated with full autonomy, seamlessly transitioning between environmental perception, AI-driven decision-making, real-time force validation, and robotic manipulation without human intervention.

The performance breakdown revealed that scene analysis and segmentation were completed in under 3 seconds, with the LLM-based site evaluation requiring approximately 2.5 seconds per frame, resulting in a total processing time of less than 18 seconds. Once the AI Space Cortex determined viable sampling locations, the system conducted force-torque probing and validation over a 230-second period, ensuring that the selected site met scoopability and material safety requirements. Following this validation, the sample retrieval and cache placement process took approximately 124 seconds, during which the robotic arm executed a controlled excavation maneuver and deposited the sample into the designated cache.

The system demonstrated full operational autonomy, transitioning seamlessly between environmental perception, AI-driven decision-making, real-time force validation, and robotic manipulation.

To evaluate the consistency and efficiency of our online calibration procedure, we conducted approximately 10 trials across diverse initial conditions. The calibration accuracy was observed to vary between 90\% and 98\%, with the number of samples required for convergence ranging from 15 to 40 sampling points. These results directly depend on solving the underlying nonconvex Bayesian optimization problem, which guides the selection of informative poses using the GP-UCB strategy. In parallel, we conducted repeated semantic reasoning tests under similar environmental conditions using our multimodal LLM-based perception system. While the system qualitatively reproduced consistent and accurate target selection behavior, its statistical quantification remains challenging due to the probabilistic and context-dependent nature of large language model outputs. Nevertheless, the qualitative reproducibility of correct sampling location identification observed across multiple trials supports the robustness of our integrated decision-making pipeline.

\section{CONCLUSIONS}
\label{sec:conclusion}
The AI Space Cortex represents a significant advancement in autonomous planetary exploration, demonstrating the capability to perceive, analyze, and interact with unknown environments in real time. Through its integration of hierarchical control, vision-based perception, LLM-driven reasoning, and fault recovery mechanisms, the system successfully executes science-driven sampling missions with minimal human intervention. We move away from traditional spacecraft autonomy approaches by incorporating generalization into the framework and allowing autonomy to adapt its approach given high-level objectives. At the same time we keep AI Space Cortex grounded in the traditional rigor required for spacecraft autonomy by building in verification steps (like cone penetrometer measurements, sample collection verification) that keep checks and balances on our autonomy and by providing interfaces for which human observers may understand and engage with the decision making process. The results presented in this study validate the AI Space Cortex’s ability to autonomously detect and correct mechanical faults, evaluate scientifically promising sampling sites, and execute complex robotic interactions. It is able to do this while maintaining explainability and operator oversight via the Explanation Engine.

We developed a Bayesian optimal experimental design method for online kinematic calibration, utilizing geometry-aware valid kernels on ${\mathbb{S}^3 \!\times\! \mathbb{R}^3}$. Instead of framing the design problem in joint space, the method optimizes directly over end-effector poses, which enhances robustness to uncertainties in joint errors and improves overall calibration accuracy. A GP-based learning method was proposed, incorporating a quaternion geodesic distance, the Euclidean distance, and a GP-UCB optimization method. Experiments conducted on a 7DOF robotic arm demonstrated the effectiveness of the method.

We executed a full sampling mission from initial perception to sample delivery, demonstrating an end-to-end execution time of 8 minutes and 40 seconds. The AI Space Cortex transitioned seamlessly between scene segmentation, scientific site evaluation via LLM, force-torque material validation, and precision robotic manipulation, ensuring that sampling operations were conducted efficiently and scientifically optimally.

The findings of this research confirm that the AI Space Cortex is a powerful tool for autonomous science missions, providing an intelligent, adaptable, and fault-resilient system for future space exploration. By combining LLM-based AI, real-time robotic execution, and structured decision transparency, the AI Space Cortex sets a foundation for the next generation of AI-driven planetary exploration architectures, enabling scientific discovery beyond Earth.

\section{FUTURE WORK}
\label{sec:future}
While we present the capabilities of AI Space Cortex in this paper, we also acknowledge its limitations with respect to application onto modern spacecraft. We present directions of expansion for our research that could directly address these limitations and incorporate lessons learned from our deployment of AI Space Cortex onto a flight-like testbed.

Spacecraft compute is limited and often shared across a variety of subsystems and tasks with changing priority. In our own deployment on a real-time hardware testbed, the computational demands of LLMs posed significant challenges for task scheduling and execution. To address these constraints, future work should explore strategies for scaling down the AI Space Cortex framework to operate effectively within the computational limits of spacecraft hardware. As large models continue to be optimized and evolve, benchmarking efforts will be essential to evaluate the performance of AI Space Cortex on next-generation analog spacecraft using emerging models such as xAI's Grok and ChatGPT on increasingly capable embedded hardware.
An additional avenue for future research lies in reimagining the architectural design of planetary missions to better balance the need for local autonomy with access to high-performance remote computation. On Earth, state-of-the-art LLM performance is typically achieved by routing inference requests through APIs to large-scale cloud infrastructure. A similar paradigm could be adapted for planetary exploration: rovers and autonomous systems with limited onboard compute capabilities could interface with a stationary lander outfitted with a compact, high-performance computing system—such as a scaled-down GPU server rack—acting as a local inference node.

This work could also be extended to benefit other types of autonomous missions; making the framework generalizable and adaptable was a focus of this work. The intrinsic adaptability of the AI Space Cortex makes it a flexible exploration framework, well-suited for deployment in a variety of planetary environments beyond icy moons. The system architecture can be integrated into different robotic platforms to support a range of extraterrestrial exploration objectives. In lunar and Martian environments, the Cortex could facilitate lava tube explorations, leveraging its ability to navigate and analyze subsurface voids using rovers or quadrupedal robotic systems. In regions with extreme topographical features, such as mountainous or canyon terrains like Tharsis Montes on Mars, the Cortex could enhance autonomous navigation and science operations for exploratory rover platforms. Additionally, the AI-driven framework could be extended to cave system investigations, utilizing aerial or unconventional robotic systems such as drones or bio-inspired robots like the Exobiology Extant Life Surveyor (EELS) \cite{vaquero2024eels} for Europa exploration. By enabling intelligent, self-directed decision-making across diverse planetary environments, the AI Space Cortex represents a scalable and adaptable solution for next-generation robotic space exploration.

\section*{ACKNOWLEDGMENT}
This work was supported by NASA Grant (80NSSC21K1032). A portion of this research was carried out at the Jet Propulsion Laboratory, California Institute of Technology, under a contract with the National Aeronautics and Space Administration (80NM0018D0004). We thank PESTO’s COLDTech program for funding this work, and the JPL Robotic Mobility team for their assistance with the OWLAT testbed and simulator. We would also like to thank the Caltech and JPL administrative teams, and in particular, Kristen Bazua.

\bibliographystyle{IEEEtran}
\bibliography{Bib/refs}

\vfill\pagebreak

\end{document}